\documentclass[11pt]{article}

\usepackage{acl}

\usepackage{times}
\usepackage{adjustbox}
\usepackage{latexsym}
\usepackage{amssymb}
\usepackage{booktabs}
\makeatletter
\providecommand{\captionof}[1]{\def\@captype{#1}\caption}
\makeatother
\usepackage{xcolor}
\usepackage{array}
\usepackage{tabularx}
\usepackage{seqsplit}
\usepackage{multirow}
\usepackage{amsmath}
\usepackage{listings}
\usepackage[most]{tcolorbox}
\usepackage[T1]{fontenc}

\usepackage[utf8]{inputenc}

\usepackage{microtype}

\usepackage{inconsolata}

\usepackage{graphicx}

\usepackage{cleveref}

\title{Evaluating the Robustness of Proof Autoformalization in Lean~4}

\author{Zhengtao Gui$^1$ \quad Sheng Yang$^2$ \quad Zhouxing Shi$^2$ \\
  $^1$University of California, Irvine \quad $^2$University of California, Riverside \\
  \texttt{zhengtag@uci.edu} \quad \texttt{syang362@ucr.edu} \quad \texttt{zhouxing.shi@ucr.edu}
}
\hypersetup{pdfauthor={Zhengtao Gui, Sheng Yang, Zhouxing Shi},pdftitle={Evaluating the Robustness of Proof Autoformalization in Lean 4},pdfsubject={},pdfkeywords={}}

\begin{document}
\maketitle
\begin{abstract}

Proof autoformalization aims to translate a mathematical informal proof in natural language into a formal proof in a formal language such as Lean~4. Existing LLM-based models for proof autoformalization are typically evaluated only on well-formed informal proofs from curated datasets. We argue that a robust proof autoformalizer must remain faithful even for informal proofs that diverge from these idealized ones, and we present the first study on the robustness of proof autoformalization models. We formulate two categories of perturbations: a global perturbation paraphrases the informal theorem--proof pair in a different style, under which the formalization should remain consistent; a local perturbation alters a value, symbol, or proof step, possibly in a counterfactual way, and a robust formalization should faithfully reflect the perturbation rather than reverting to the original or inferring a different one on its own. We build a benchmark with both perturbations on miniF2F and MATH-500, and automatically measure how stable a proof autoformalization's correctness is under global perturbations and how faithfully its output reflects local perturbations. We evaluate seven recent models: their correctness varies under global perturbations, and they mostly fail to remain faithful under local perturbations. Code and data are available at \url{https://github.com/ucr-rai/robust-proof-autoformalization}.
\end{abstract}

\begin{figure*}[t]
    \centering
    \includegraphics[width=\textwidth]{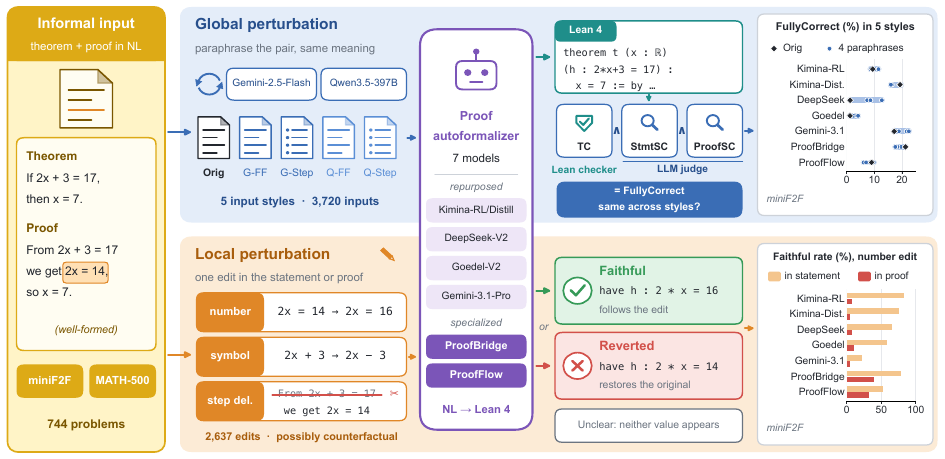}
    \caption{Overview of our study. Each informal theorem--proof pair is perturbed in two ways and autoformalized by seven models. \textbf{Top:} global perturbations paraphrase the pair into four styles, and we check whether FullyCorrect stays the same. \textbf{Bottom:} local perturbations edit one number, symbol, or step, and each output is labeled faithful, reverted, or unclear. \textbf{Right:} on miniF2F, correctness varies with style, and edits in the proof are rarely followed.}
    \label{fig:overview}
\end{figure*}

\section{Introduction}

Large language models (LLMs) have recently demonstrated strong capabilities in mathematical reasoning~\citep{lightman2024let,guo2025deepseek}, which is typically conducted in natural language; however, such natural-language reasoning is inherently challenging to verify, and it is hard to detect subtle logical errors reliably. In contrast, formal theorem provers, such as Coq~\cite{barras1999coq}, Isabelle~\cite{nipkow2002isabelle},
and Lean 4~\cite{moura2021lean}, represent mathematical statements and proofs in a rigorous formal language, enabling  formal verification of whether a proof is correct for a given statement.

Given the strong capabilities of LLM reasoning developed in the natural-language (NL) space and the formal reliability provided by the formal-language (FL) space, bridging NL and FL reasoning has emerged as an active research direction. The bridge can run in both directions: NL reasoning can guide the generation of FL proofs~\citep{jiang2022draft,ren2025deepseek,wang2025kimina}, while FL verification can in turn verify NL reasoning once it is faithfully translated into FL~\citep{liu2025safe,jana2025proofbridge,cabral2025proofflow}.

The translation from NL mathematics into FL is known as \emph{autoformalization}, which can be categorized into statement autoformalization and proof autoformalization, depending on whether only the statement or also the proof is translated. Most of the previous work has focused on statement autoformalization~\cite{gao2024herald,jiang2024multi,poiroux2025reliable,wang2025kimina,wu2026stepfun}.
Proof autoformalization, which is substantially more challenging, has gained attention only recently~\citep{jana2025proofbridge,cabral2025proofflow} and is the focus of this work.
However, these methods are typically only trained and tested on well-formed proofs from standard datasets, and it remains unclear how robust they are when the input deviates from idealized proofs.

In this work, we present the first study on the robustness of proof autoformalization. Specifically, we evaluate robustness under two categories of perturbations applied to the NL input (\Cref{fig:overview}).
First, when the NL proof is correct but exhibits stylistic variations, the resulting FL proof should remain correct. This robustness is crucial for a proof autoformalization model to remain effective in practice across NL proofs of varying styles and quality.  We thus design \emph{global perturbations}, which paraphrase the entire NL theorem--proof pair in different styles while preserving the mathematical semantics.
Second, when a subtle change is introduced into the NL proof --- even a counterfactual one that represents an error --- the autoformalization should faithfully reflect this change in the FL proof rather than inferring a different one on its own, for instance by silently correcting the error. Indeed, proving a theorem and faithfully translating its supplied informal proof are distinct objectives.
This robustness provides a basis for future work that may combine autoformalization with Lean-based formal verification to verify NL proofs and detect errors in them, so that NL reasoning can in turn be strengthened by feedback from the FL side.
We therefore design \emph{local perturbations}, which modify a single value, symbol, or proof step in the NL proof, possibly in a counterfactual way.

We summarize our contributions as follows:
\begin{itemize}
    \item We present the first study on the robustness of proof autoformalization, designing global and local perturbations with accompanying robustness metrics to assess properties that a robust autoformalization model should satisfy.
    \item We build a benchmark that applies these perturbations to miniF2F~\citep{zheng2021minif2f} and MATH-500~\citep{lightman2024let}, and evaluate seven models on it, including recent proof autoformalization methods such as ProofBridge~\citep{jana2025proofbridge} and ProofFlow~\citep{cabral2025proofflow}.
    \item Our experiments show that all seven models have limited robustness: their correctness is unstable when a correct proof is paraphrased, and they often restore the original content instead of faithfully translating a local edit, indicating substantial room for improvement. These findings are supported by reliability checks of the perturbations and the LLM judges, and by additional analyses on proof length and on multiple sampled outputs.
\end{itemize}

\section{Related Work}

\paragraph{Formal Theorem Proving.}
Formal theorem proving generates verifiable FL proofs from FL statements. Existing methods generally follow two paradigms. Next-tactic prediction generates one tactic at a time from the current proof state, as in GPT-f~\cite{polu2020generative}, LISA~\cite{jiang2021lisa}, and PACT~\cite{han2021proof}. Whole-proof generation produces a complete proof script, as in DeepSeek-Prover-V1~\cite{xin2024deepseek}, TheoremLlama~\cite{wang2024theoremllama}, DeepSeek-Prover-V2~\cite{ren2025deepseek}, Kimina-Prover~\cite{wang2025kimina}, and Goedel-Prover-V2~\cite{lin2025goedelv2}. Verifier feedback also supports reinforcement learning, as in Kimina-Prover~\cite{wang2025kimina}.
The primary focus of all these methods is on generating correct FL proofs, while we focus on bridging NL and FL proofs so that the generated FL proof faithfully reflects the given NL proof.

\paragraph{Autoformalization.}
Autoformalization translates informal statements or proofs into FL. Statement autoformalization is more established and expresses an informal problem's hypotheses and conclusion as a formal proposition. Recent LLM-based approaches use in-context learning~\cite{wu2022autoformalization}, back-translation-based data synthesis~\cite{jiang2024multi,azerbayev2023proofnet}, retrieval~\cite{liu2025rethinking}, natural language inference~\cite{ying2024lean}, judge-assisted expert iteration~\cite{wang2025kimina}, and translator fine-tuning on constructed datasets~\cite{gao2024herald}. Back-translation renders formal statements into NL to construct paired training examples~\cite{jiang2024multi}.

Proof autoformalization has received relatively limited attention and is a more recent research direction.
FormL4~\cite{lu2024process} uses GPT-4-informalized Mathlib proofs and step-wise Lean feedback; ProofBridge~\cite{jana2025proofbridge} uses NL-FL representation learning and cross-modal retrieval to guide formalization. Step-wise methods include StepProof in Isabelle~\cite{hu2025stepproof} and ProofFlow~\cite{cabral2025proofflow}, which represents proof-step dependencies with a DAG and formalizes intermediate lemmas.
Related studies examine the robustness of statement autoformalization to paraphrases~\cite{moore2025evaluating,feng2026paraphrasefailures}.
In this work, we focus on evaluating the robustness of existing proof autoformalization methods in Lean~4.

\paragraph{Metrics for autoformalization.}
For proof autoformalization, recent evaluations introduce proof-level metrics.
ProofBridge~\cite{jana2025proofbridge} reports Type Correctness (TC), determined by Lean's type checker, and Semantic Correctness (SC), measuring equivalence to gold FL only for type-correct outputs. ProofFlow's \emph{proofscore} combines step-level type checking, semantic faithfulness, and DAG-based structural fidelity~\cite{cabral2025proofflow}.
These metrics, typically evaluated on well-formed NL proofs, do not assess the robustness of proof autoformalization, which we investigate in this work.

\section{Method}
\label{sec:method}

We study the robustness of proof autoformalization models in Lean~4.
For a target model $f$, the task is to translate a natural-language (NL) theorem--proof pair $M_{\mathrm{NL}}=\langle T_{\mathrm{NL}}, P_{\mathrm{NL}}\rangle$ into a formal-language (FL) pair $\widehat{M}_{\mathrm{FL}}=\langle \widehat{T}_{\mathrm{FL}}, \widehat{P}_{\mathrm{FL}}\rangle$:
\[ \widehat{M}_{\mathrm{FL}} = f(M_{\mathrm{NL}}),\]
where $T_{\mathrm{NL}}$ and $P_{\mathrm{NL}}$ denote the NL theorem statement and proof, and $\widehat{T}_{\mathrm{FL}}$ and $\widehat{P}_{\mathrm{FL}}$ denote their FL counterparts.
We investigate the robustness under two categories of perturbations applied to the NL input.
A \emph{global} perturbation paraphrases the whole NL theorem--proof pair $M_{\mathrm{NL}}$, and we evaluate robustness to it by measuring the correctness of the resulting formalization (\Cref{sec:global-perturbation}); a \emph{local} perturbation changes a single value, symbol, or proof step within the NL input, and we evaluate faithfulness by measuring whether the formalization faithfully reflects the change (\Cref{sec:local-perturbation}).

\subsection{Global Perturbations}
\label{sec:global-perturbation}
For global perturbations, we use LLMs to paraphrase both the NL theorem statement and its proof.
Specifically,
we use two LLMs: Gemini-2.5-Flash~\citep{google2025gemini25flash} (\textbf{G}), representing a closed-source model, and Qwen3.5-397B-A17B~\citep{qwen35ModelCard} (\textbf{Q}), representing an open-source model.
Given the original NL theorem--proof pair, denoted as \textbf{Orig}, we consider two paraphrasing modes:
\emph{free-form paraphrasing}~(\textbf{-FF}) uses an LLM to restate the pair in its own wording and style while preserving the mathematical meaning and overall logic;
and \emph{step-by-step paraphrasing}~(\textbf{-Step}) restates the statement and reorganizes the proof into a clearly numbered sequence of steps.
For example, ``Since $x^2-5x+6=(x-2)(x-3)$, the roots are $2$ and $3$'' can be restated freely as ``Factoring $x^2-5x+6$ gives $(x-2)(x-3)$, yielding roots $2$ and $3$'', or reorganized into numbered factoring and root-identification steps. The mathematical claim is unchanged.
Using Gemini-2.5-Flash and Qwen3.5-397B-A17B, respectively, we generate four types of globally perturbed NL theorem--proof pairs:
\textbf{G-FF}, \textbf{Q-FF}, \textbf{G-Step}, and \textbf{Q-Step}.
Together with \textbf{Orig}, this yields five NL input styles per problem. Details are in \Cref{app:perturbation-construction}.

\paragraph{Correctness Metrics.}
\label{sec:correctness-metrics}
Since these global perturbations are expected to preserve the mathematical meanings, a robust proof autoformalization should achieve comparable correctness across all five styles.
We therefore measure \emph{robustness} by the stability of correctness across styles: large variation across styles indicates sensitivity to surface wording rather than the underlying mathematics.
As shown in \Cref{fig:overview} (top), each generated FL output
is evaluated by three complementary correctness metrics:
\textbf{Type Correctness (TC)} checks whether the generated FL output passes the Lean type checker without errors. TC accepts outputs containing \texttt{sorry}, which Lean reports as a warning rather than an error; such outputs are rare among the FullyCorrect ones we inspect (\Cref{app:full-results}).
\textbf{Statement Semantic Correctness (StmtSC)} and \textbf{Proof Semantic Correctness (ProofSC)} assess whether the generated formal statement and proof are semantically consistent with their respective NL counterparts, i.e., the NL theorem and proof given to the model as input.
StmtSC and ProofSC are assigned by an LLM judge (detailed in \Cref{app:sc-judge-prompts}).
\textbf{FullyCorrect} requires TC, StmtSC, and ProofSC to all pass.
Judging whether given NL and FL texts are semantically consistent is considerably easier than generating the FL output, so we expect the judge to be much more stable across input styles than the evaluated models, which we further support with a second judge in \Cref{sec:reliability-data}.

\subsection{Local Perturbations}
\label{sec:local-perturbation}

A local perturbation changes a single value, symbol, or proof step in the
NL input. We deliberately allow such perturbations to be possibly counterfactual, in order to reveal the different behaviors of two types of autoformalizers.
A \textbf{Faithful Autoformalizer} produces a Lean formalization that faithfully reflects the NL input it is given, even in the presence of counterfactual edits, which is the desired behavior for robust autoformalization.
In contrast, a \textbf{Biased Autoformalizer} produces an autoformalization that silently overrides the edit with its own inference.
This occurs because a spurious bias acquired during training may lead the model to fill in seemingly plausible content based on the FL statement, the context of the FL proof, and patterns seen in well-formed proofs, rather than faithfully following the actual NL input.
For example, given a proof
``from $2x+3=17$ we get $2x=14$, so $x=7$'', we may change the intermediate step to $2x=16$, which no longer holds arithmetically.
A Faithful Autoformalizer should carry this step into the FL proof and produce \verb|2 * x = 16|, whereas a Biased Autoformalizer may revert to 14 on its own and produce \verb|2 * x = 14|.

We design three specific types of edits for the local perturbations.
A \textbf{number edit} changes a numeric literal to a
different value of similar magnitude and the same sign; a \textbf{symbol edit}
replaces a relation or operator with its opposite (e.g.\ $+$ with $-$, or $\le$
with $\ge$); and a \textbf{step deletion} removes the justification of one
substantive proof step while keeping its stated outcome, so the proof asserts
the step's result without proving it.
For example, consider a proof step that derives a lower bound:
\begin{quote}
Since $(x-3)^2\ge 0$, expanding gives $x^2-6x+9\ge 0$. Hence
$x^2-6x+10\ge 1$.
\end{quote}
A number edit may change the final bound from $1$ to $2$. A symbol edit may
change the relation $x^2-6x+10\ge 1$ to $x^2-6x+10\le 1$, or change the
operator in $x^2-6x+10$ from $+$ to $-$. A step deletion may remove the
justification ``Since $(x-3)^2\ge 0$, expanding gives
$x^2-6x+9\ge 0$'' while keeping the stated outcome
``Hence $x^2-6x+10\ge 1$.''
In all three cases, the edited proof may no longer
be mathematically valid, but a faithful formalization should still reflect the edited text rather than silently repairing it. This is particularly important when proof autoformalization is applied to verify an NL proof---possibly incorrect---by checking its autoformalized counterpart in a formal proof assistant such as Lean.

Number and symbol edits are applied in both the proof and the statement; step deletion applies only to the proof. The statement edits act as a reference against which the proof-level results can be read. To build an edit, an LLM (Gemini-2.5-Flash) first marks the eligible locations in the input: numeric literals, relations and operators, and substantive proof steps. We keep only locations that pass rule-based validity checks, discarding cases that would render the problem malformed, such as numbers inside subscripts or superscripts and relations inside chained expressions like $0<p<15$. From the valid locations, we choose one for each of the statement and the proof. For a number or symbol edit, this is a single-span substitution at a marked position; for a step deletion, we instead select one eligible step and delete its justification. \Cref{app:perturbation-construction} gives the full marking and filtering rules, and \Cref{tab:edit-counts} reports the per-type edit counts.

\paragraph{Faithfulness Metrics.}
\label{sec:faithfulness-metrics}
For local perturbations, we classify each output as faithful, reverted, or unclear (\Cref{fig:overview}, bottom): a \emph{faithful} output reflects the edited input, a \emph{reverted} output matches the original unedited input, and an \emph{unclear} output matches neither (for example, it is empty or never mentions the edited entity). For number and symbol edits, this label is an objective property of the output rather than a subjective judgment: it is determined entirely by which value the output carries (the edited one, the original, or neither).
The labels do not depend on whether the edited input still constitutes correct reasoning; in either case, a faithful output reflects the edited NL input as given.
For step deletion, a faithful output retains the gap left by the deletion: the claim whose justification was removed is left unproved or simply assumed, rather than re-derived. We only delete reasoning that the original proof states explicitly, such as applying a theorem or carrying out a calculation. The gap is therefore a missing part of the argument, not a routine detail that NL proofs usually leave to the formalizer. Filling in the gap may help prove the theorem, but a faithful autoformalization should keep the gap visible in the FL proof, so that the incompleteness of the NL proof can be exposed.
We thus consider the gap retained when the FL output leaves the claim to \texttt{sorry}, \texttt{admit}, an unfilled metavariable, or an empty proof body, or uses the claim without proving it. If the output instead supplies tactics that derive the claim, it reconstructs the deleted reasoning and is labeled reverted, even if the derivation is a single short tactic; if the claim cannot be located in the output, the output is labeled unclear. \Cref{app:edit-judge-prompts} gives the judge prompt.
We assign these labels with an LLM judge; such labels can be easily inspected by humans, and we evaluate their reliability in \Cref{app:reliability-detail}. The labels are assigned regardless of whether the output passes TC, since an FL output can reflect the edit even if it does not type-check.
From these labels we compute the \textbf{Faithful Rate (FR)} and \textbf{Reverted Rate (RR)} as the fractions of the identifiable (faithful or reverted) outputs, so that $\mathrm{FR}+\mathrm{RR}=1$; a higher FR marks a Faithful Autoformalizer and a higher RR indicates a Biased Autoformalizer. The unclear outputs are excluded from FR/RR, and we report their share as the \textbf{Unclear Rate (UR)}, analyzing its composition with examples in \Cref{app:other-breakdown}.

\section{Experiments}
\label{sec:experiments}

We first describe the common experimental setup (\Cref{sec:setup}), then
report two experiments: one measuring robustness to global perturbations
through the stability of correctness (\Cref{sec:exp-robustness}), and one
measuring faithfulness to local perturbations (\Cref{sec:exp-faithfulness}). Finally, we validate the perturbations and the LLM judges (\Cref{sec:reliability-data}).

\subsection{Setup}
\label{sec:setup}

\paragraph{Datasets.}
\label{sec:datasets}
We build our evaluation based on two mathematical reasoning datasets: miniF2F~\cite{zheng2021minif2f} and MATH-500~\cite{lightman2024let}. We use the 244-problem test split of miniF2F, whose full benchmark contains 488 problems drawn from competitions such as AMC, AIME, and IMO and from high-school and undergraduate courses, and the 500 problems of MATH-500 drawn from MATH~\cite{hendrycks2021math}. MATH-500 problems are originally posed as computational problems that require producing concrete answers;
since a Lean statement must be a theorem that requires a proof rather than a computational answer,
we append each gold answer to its statement as a ``Show that it is $X$'' clause, turning the problem into a closed proposition requiring a proof, as similarly done in prior work using Lean for answer selection~\citep{yao-etal-2025-fans}. The informal statements and proofs of miniF2F are those released by \citet{jiang2022draft}, and for MATH-500 the reference solution serves as the informal proof.

For global perturbations,
each of the problems appears in five styles, which gives 3,720 problem-style instances in total.
For local perturbations, we generate one edit per problem for each applicable
combination of edit type and location: number and symbol edits in both the
statement and the proof, and step deletion in the proof only. An edit is
created only where an eligible location exists, and thus the counts vary by type and
dataset.
\Cref{tab:edit-counts} shows the full breakdown, totaling
2,637 local perturbations. Our code and data are available at \url{https://github.com/ucr-rai/robust-proof-autoformalization}.

\paragraph{Models.}
\label{sec:models}
We evaluate seven models in total, which we organize into two groups. \textbf{(i) Repurposed provers and a general-purpose LLM.} Kimina-Prover-RL-1.7B and Kimina-Prover-Distill-8B~\cite{wang2025kimina}, DeepSeek-Prover-V2-7B~\cite{ren2025deepseek}, and Goedel-Prover-V2-8B~\cite{lin2025goedelv2} are specialized theorem provers in Lean~4, and they are trained to generate a formal proof for a given formal statement. Although these are specialized models for proof generation rather than autoformalization, given the scarcity of proof autoformalization models, we follow \citet{jana2025proofbridge} and repurpose these models for proof autoformalization with new prompts. Their scores here are therefore not comparable to their results on theorem-proving benchmarks, which start from a curated formal statement and allow many attempts. Additionally, we adopt Gemini-3.1-Pro~\citep{google2026gemini31pro}, which is a general-purpose LLM and can thus be prompted to perform proof autoformalization.
\textbf{(ii) Specialized proof autoformalization models.} ProofBridge and ProofFlow are the two models in our evaluation built specifically for proof autoformalization, in contrast to the repurposed provers and the general-purpose LLM above.
\textbf{ProofBridge}~\citep{jana2025proofbridge}
is trained by Supervised Fine-tuning (SFT) from Kimina-Prover-RL-1.7B on proof autoformalization data. Although ProofBridge is enhanced with a retrieval-augmented approach which guides the autoformalization with similar cases from the database, this component has not been included in their released code; we thus focus on the SFT model, which we train with their released code. Nevertheless, we do not expect the retrieval-augmented approach to improve the robustness we study, as it uses extra references from the database rather than the current NL input.
\textbf{ProofFlow}~\citep{cabral2025proofflow} is a formalization pipeline designed to preserve proof structure. It constructs a directed acyclic graph (DAG) of proof-step dependencies, formalizes intermediate lemmas, and then attempts to complete the proof through these lemmas.
We use the decoding settings in \Cref{app:experimental-setup} and evaluate one output per input for every model; \Cref{app:additional-results} additionally studies multiple sampled outputs.

All semantic and faithfulness metrics (StmtSC, ProofSC, FR/RR/UR) are assigned by Gemini-2.5-Flash, so that cross-model comparisons are not confounded by judge differences. To rule out a bias specific to this judge, we further validate the semantic labels with a second judge (\Cref{sec:reliability-data}).
Type checking uses Lean~4~\cite{moura2021lean}
with a $900$-second per-problem timeout. Details are documented in \Cref{app:implementation-detail}.

\subsection{Robustness to Global Perturbations}
\label{sec:exp-robustness}

\begin{figure}[t]
    \centering
    \includegraphics[width=\columnwidth]{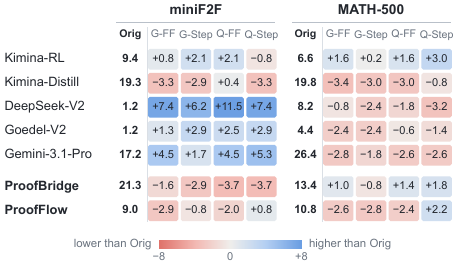}
    \caption{FullyCorrect (\%) on the original input (Orig) and its change in percentage points on each paraphrased style. An output is FullyCorrect only if it passes TC, StmtSC, and ProofSC. Blue cells are higher than Orig and red cells are lower; \Cref{tab:robustness} gives the absolute values.}
    \label{fig:delta-fullycorrect}
\end{figure}

\Cref{fig:delta-fullycorrect} reports FullyCorrect on the original input and its change under four global perturbations, each paraphrasing both the theorem statement and the proof. A robust autoformalizer should achieve comparable FullyCorrect across columns, but FullyCorrect changes with the NL input style for every model. On miniF2F, the change is largest for DeepSeek-Prover-V2-7B, which rises from $1.2\%$ on Orig to $12.7\%$ on Q-FF, while ProofBridge drops from $21.3\%$ on Orig to $17.6\%$ under both Q-FF and Q-Step. Despite being the strongest model in aggregate on both datasets and presumably much larger than the other models, Gemini-3.1-Pro still ranges from $17.2\%$ to $22.5\%$ across styles on miniF2F.

Paraphrasing has mixed effects across models and datasets.
On miniF2F, paraphrasing can improve some models: besides DeepSeek-Prover-V2-7B above, Goedel-Prover-V2-8B rises from $1.2\%$ on Orig to $4.1\%$ on G-Step and Q-Step. On MATH-500, however, paraphrasing often lowers performance: for example, Gemini-3.1-Pro drops from $26.4\%$ to $23.6$--$24.6\%$.
Thus, its effect depends on both the dataset and the model.

Overall, no input style dominates uniformly, and semantic-preserving
paraphrases can change both absolute scores and model rankings.
For example, on miniF2F, ProofBridge leads Gemini-3.1-Pro by $4.1$ points on Orig, whereas Gemini-3.1-Pro leads on all four paraphrased styles, by $0.5$ to $4.9$ points. This indicates that current autoformalizers are not stable under global perturbations with paraphrased NL inputs, and that evaluations solely on a fixed original style can over- or under-state a model's true autoformalization ability.

We also find that, for ProofBridge and ProofFlow, the sensitivity to styles exists for both shorter and longer proofs (\Cref{tab:additional-proof-length}). Moreover, allowing ProofBridge to select the first type-checked proof across multiple sampled proofs does not eliminate the sensitivity (\Cref{tab:additional-verifier-selection}). With eight outputs per input, the style has a small but consistent effect beyond sampling (\Cref{app:additional-results}).

\begin{figure*}[t]
    \centering
    \includegraphics[width=\textwidth]{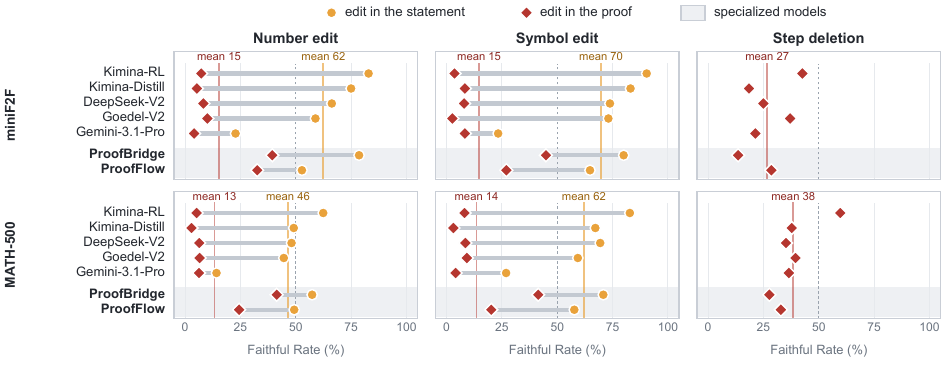}
    \caption{Faithful Rate (FR, \%) under local perturbations, from \Cref{tab:proof-edit-results,tab:statement-edit-results}. For number and symbol edits, each line connects the FR of one model when the edit is in the statement (circle) and when it is in the proof (diamond); step deletion applies only to the proof. The dotted line marks $50\%$, the vertical colored lines mark the mean over the seven models, and the shaded band marks the two specialized models.}
    \label{fig:faithful-rate}
\end{figure*}

\subsection{Faithfulness under Local Perturbations}
\label{sec:exp-faithfulness}

\label{sec:faithfulness-results}

\begin{table*}[!t]
\centering

\setlength{\tabcolsep}{3pt}
\renewcommand{\arraystretch}{1.05}
\adjustbox{max width=.99\textwidth}{\begin{tabular}{l|cc|cc|cc|cc|cc|cc}
\toprule
\multirow{3}{*}{\textbf{Model}}
& \multicolumn{4}{c|}{\textbf{Number edit}}
& \multicolumn{4}{c|}{\textbf{Symbol edit}}
& \multicolumn{4}{c}{\textbf{Step deletion}} \\
\cmidrule(lr){2-5} \cmidrule(lr){6-9} \cmidrule(lr){10-13}
& \multicolumn{2}{c|}{miniF2F} & \multicolumn{2}{c|}{MATH-500}
& \multicolumn{2}{c|}{miniF2F} & \multicolumn{2}{c|}{MATH-500}
& \multicolumn{2}{c|}{miniF2F} & \multicolumn{2}{c}{MATH-500} \\
\cmidrule(lr){2-3} \cmidrule(lr){4-5}
\cmidrule(lr){6-7} \cmidrule(lr){8-9}
\cmidrule(lr){10-11} \cmidrule(lr){12-13}
& FR ($\uparrow$) & RR ($\downarrow$)
& FR ($\uparrow$) & RR ($\downarrow$)
& FR ($\uparrow$) & RR ($\downarrow$)
& FR ($\uparrow$) & RR ($\downarrow$)
& FR ($\uparrow$) & RR ($\downarrow$)
& FR ($\uparrow$) & RR ($\downarrow$)
\\
\midrule
Kimina-Prover-RL-1.7B
& 7.3 & 92.7
& 5.2 & 94.8
& 3.6 & 96.4
& 8.1 & 91.9
& 42.6 & 57.4
& 59.7 & 40.3 \\
Kimina-Prover-Distill-8B
& 5.3 & 94.7
& 2.9 & 97.1
& 8.3 & 91.7
& 3.1 & 96.9
& 18.5 & 81.5
& 37.8 & 62.2 \\
DeepSeek-Prover-V2-7B
& 8.3 & 91.7
& 6.4 & 93.6
& 7.9 & 92.1
& 8.5 & 91.5
& 25.0 & 75.0
& 35.2 & 64.8 \\
Goedel-Prover-V2-8B
& 10.1 & 89.9
& 6.6 & 93.4
& 2.6 & 97.4
& 9.2 & 90.8
& 37.1 & 62.9
& 39.5 & 60.5 \\
\midrule
Gemini-3.1-Pro
& 4.1 & 95.9
& 6.3 & 93.7
& 8.3 & 91.7
& 4.1 & 95.9
& 21.4 & 78.6
& 36.5 & 63.5 \\
\midrule
ProofBridge
& 39.4 & {60.6}
& {41.4} & {58.6}
& {44.9} & {55.1}
& {41.4} & {58.6}
& 13.6 & 86.4
& 27.7 & 72.3 \\
ProofFlow
& 32.6 & 67.4
& 24.4 & 75.6
& 27.0 & 73.0
& 20.2 & 79.8
& 28.6 & 71.4
& 32.9 & 67.1 \\
\bottomrule
\end{tabular}}
\caption{Faithful Rate (FR) and Reverted Rate (RR) (\%) for proof-level local perturbations, computed over identifiable outputs (faithful or reverted). Unclear outputs are excluded, and the identifiable fraction is reported in \Cref{tab:identifiable-fraction}. Models are grouped as repurposed provers (top), a general-purpose LLM (middle), and specialized proof autoformalization models (bottom).}
\label{tab:proof-edit-results}
\end{table*}

We use the three local perturbations of \Cref{sec:local-perturbation}, each testing a specific faithful behavior. \Cref{tab:proof-edit-results,tab:statement-edit-results} summarize FR (the higher the better) and RR (the lower the better) across proof-level and statement-level local perturbations, respectively, and \Cref{fig:faithful-rate} plots FR for both levels.

Models are least faithful on proof-level number and symbol edits. For the five models other than ProofBridge and ProofFlow, the average FR on these edits is only $6.3\%$. RR is above $90\%$ in almost all corresponding cells. For example, on miniF2F number-proof edits, Kimina-Prover-Distill-8B has $94.7\%$ RR and Gemini-3.1-Pro has $95.9\%$ RR. Counting unclear outputs as well, these five models follow a proof-level number edit in fewer than $6\%$ of all outputs (\Cref{fig:edit-outcomes}).
ProofBridge and ProofFlow, which are specialized proof autoformalization models, are relatively more faithful on proof-level number and symbol edits. ProofBridge has $41.8\%$ average FR on these edits, and ProofFlow has
$26.1\%$, both higher than the $6.3\%$ average of the other five models. However, this advantage is limited: even for these proof-level edits, both systems still revert in more than half of identifiable cases. For instance, in one MATH-500 problem, the edited proof says ``Rewrite $4$ as $5^2$'' (originally $2^2$), yet ProofBridge still outputs \texttt{rw [show 4 = 2 \textasciicircum{} 2 by norm\_num]}.
Such reversion is consistent with the bias described in \Cref{sec:local-perturbation}. A proof-level edit is only a local change, with the statement and most of the proof context unchanged. The input overall resembles a well-formed proof, and thus a Biased Autoformalizer tends to produce a plausible FL proof that is consistent with the statement and the proof context, while silently overriding the local edit.

Under step deletion, models also tend to fill in the removed reasoning on their own.
On miniF2F, step-deletion RR ranges from $57.4\%$ to $86.4\%$, and on MATH-500 it is at least $60.5\%$ for every model except Kimina-Prover-RL-1.7B.
Kimina-Prover-RL-1.7B has the highest FR on step deletion, with $42.6\%$ on miniF2F and $59.7\%$ on MATH-500, but about two thirds of its faithful outputs simply leave the proof body empty; without them, its FR is $12.9\%$ and $39.6\%$, and no model exceeds $40\%$.
This likely reflects its limited ability to produce a proof owing to its smaller size rather than genuinely better faithfulness.
Between the two specialized models, ProofFlow retains the gap more often than ProofBridge ($28.6\%$ vs.\ $13.6\%$ FR on miniF2F).

Statement edits are much easier for models to follow than proof edits (\Cref{fig:faithful-rate}). For
every model, FR is higher when the same type of edit is made in the statement
rather than in the proof. For example, ProofBridge goes from
$39.4\%$/$44.9\%$ FR on miniF2F proof-level number/symbol edits to
$78.6\%$/$80.0\%$ on statement-level edits. Gemini-3.1-Pro also rises from
$4.1\%$/$8.3\%$ to $22.8\%$/$23.3\%$.
This is consistent with the bias described above: since a Biased Autoformalizer tends to produce an FL proof consistent with the statement, an edit in the statement is naturally more likely to be reflected in the output. Gemini-3.1-Pro is an exception: even for statement edits, it restores the original content in most identifiable outputs (RR between $73.1\%$ and $85.8\%$), consistent with a strong tendency to recall or re-derive the original problem.

\subsection{Perturbation and Evaluation Reliability}
\label{sec:reliability-data}

To ensure the validity of the results under constructed perturbations, we check the quality of the perturbations. A local edit should change only the targeted number, symbol, or step, leaving the rest of the input unchanged. We verify this with an automated script that compares every edited input against its original character by character: in all 2,637 cases the two differ in exactly one contiguous span (the intended edit) and are otherwise identical ($100\%$).
For global perturbations, an automated check finds that $96.4\%$ of all 2,976 perturbed inputs preserve every numeric and symbolic token of the original, and the remaining $3.6\%$ differ only in formatting (e.g., digit grouping or numbers spelled out) rather than content.
We also manually inspect a stratified sample of 100 pairs of original and perturbed inputs, covering the four perturbed input styles on both datasets, and find that each global perturbation changes only the wording while preserving all values, claims, reasoning steps, and semantic meaning.
The detailed procedure is given in \Cref{app:reliability-detail}, where we also evaluate the judge.

To rule out a bias specific to the Gemini-2.5-Flash judge, we randomly sample 200 FL outputs from the global perturbation experiments, 20 for each of the five input styles  on each of miniF2F and MATH-500, and re-judge their StmtSC and ProofSC with GPT-5.4-mini~\citep{openai2026gpt54mini} using the same prompts. As shown in \Cref{tab:additional-cross-judge}, the two judges give similar positive rates and agree on $88.5\%$ of StmtSC and $84.5\%$ of ProofSC labels, with Cohen's $\kappa$ of $0.735$ and $0.690$, respectively. A manual check of 100 verdicts per metric, balanced between accepted and rejected ones, also agrees with the judge on 94 StmtSC and 90 ProofSC cases. For the faithfulness labels, the judge's verdict matches the value it reports from the output in $96.7\%$ of the 16,212 number and symbol cases (\Cref{app:reliability-detail}).

\section{Conclusion}
\label{sec:conclusion}
This work presents the first study of the robustness of proof autoformalization under global paraphrases and local edits. Our results on seven models reveal two types of robustness failures: under global perturbations, the models are unstable when exposed to semantic-preserving changes in NL proof inputs; under local perturbations, they often fail to faithfully reflect local edits, and even the two models built for this task revert in more than half of the identifiable outputs under proof-level edits.
These failures indicate that current models tend to prove the theorem in their own way rather than translate the given proof; such a model cannot expose errors in an NL proof.
Our findings suggest that proof autoformalization should be evaluated beyond correctness on well-formed proofs from standard datasets; robustness to NL proofs of varying styles and quality should be considered in future development to build reliable proof autoformalizers for broader applications. Promising directions include training on paraphrased and locally edited proofs, objectives that reward consistency and faithfulness, and agentic pipelines that check faithfulness before returning a formalization.

\section*{Acknowledgement}
This work is supported in part by an NVIDIA Academic Grant award.

This work used the Delta system at the National Center for Supercomputing Applications [award OAC 2005572] through allocation CIS250704 from the Advanced Cyberinfrastructure Coordination Ecosystem: Services \& Support (ACCESS) program, which is supported by National Science Foundation grants \#2138259, \#2138286, \#2138307, \#2137603, and \#2138296.

This work used resources available through the National Research Platform (NRP) at the
University of California, San Diego. NRP has been developed, and is supported in part, by funding
from National Science Foundation, from awards 1730158, 1540112, 1541349, 1826967,
2112167, 2100237, and 2120019, as well as additional funding from community partners.

\section*{Limitations}
\label{sec:limitations}

Our study has several limitations that suggest directions for future work. First, the evaluated datasets consist of mathematical competition problems and undergraduate-level mathematics, which provide a controlled setting; as proof autoformalization models improve, harder benchmarks with longer proofs, such as PutnamBench~\citep{tsoukalas2024putnambench}, are a natural next step. Second, we target only the Lean~4 proof assistant; extending the benchmark to other systems such as Isabelle is also valuable. Third, our study evaluates existing models rather than proposing a new one. Each model is run in the configuration available to us and with one output per input: ProofBridge is the SFT model trained with its released code, without the retrieval component, and ProofFlow uses the component models listed in \Cref{app:experimental-setup}. Differences across models therefore also reflect their size and training data. Fourth, our semantic and faithfulness labels come from an LLM judge, which we validate with a second judge, deterministic checks, and manual inspection; larger-scale human annotation would further strengthen them. Fifth, our prompts ask for compilable Lean, as is standard for these models; prompts or training objectives that explicitly reward preserving gaps and errors may improve faithfulness. Lastly, a wider range of local edits and paraphrasing styles could be explored, including perturbations generated adversarially with respect to the evaluated models. We leave these extensions to future work.

\bibliography{custom,papers}

\begin{thebibliography}{43}
\providecommand{\natexlab}[1]{#1}

\bibitem[{Azerbayev et~al.(2023)Azerbayev, Piotrowski, Schoelkopf, Ayers, Radev, and Avigad}]{azerbayev2023proofnet}
Zhangir Azerbayev, Bartosz Piotrowski, Hailey Schoelkopf, Edward~W Ayers, Dragomir Radev, and Jeremy Avigad. 2023.
\newblock {ProofNet}: Autoformalizing and formally proving undergraduate-level mathematics.
\newblock \emph{arXiv preprint arXiv:2302.12433}.

\bibitem[{Barras et~al.(1999)Barras, Boutin, Cornes, Courant, Coscoy, Delahaye, de~Rauglaudre, Filli{\^a}tre, Gim{\'e}nez, Herbelin et~al.}]{barras1999coq}
Bruno Barras, Samuel Boutin, Cristina Cornes, Judica{\"e}l Courant, Yann Coscoy, David Delahaye, Daniel de~Rauglaudre, Jean-Christophe Filli{\^a}tre, Eduardo Gim{\'e}nez, Hugo Herbelin, et~al. 1999.
\newblock \href {https://pauillac.inria.fr/~herbelin/tmp/coq/V6/V6.3/doc/www/cover.html} {\emph{The {Coq} Proof Assistant Reference Manual}}.
\newblock INRIA.
\newblock Version 6.3.1, December 1999; corrected May 2000.

\bibitem[{Cabral et~al.(2026)Cabral, Do, Yu, Tai, Feng, and Shen}]{cabral2025proofflow}
Rafael Cabral, Tuan~Manh Do, Xuejun Yu, Wai~Ming Tai, Zijin Feng, and Xin Shen. 2026.
\newblock \href {https://proceedings.iclr.cc/paper_files/paper/2026/file/5fce5198dbf92d5ff45f74504431e2ff-Paper-Conference.pdf} {{ProofFlow}: A dependency graph approach to faithful proof autoformalization}.
\newblock In \emph{International Conference on Learning Representations}.

\bibitem[{Cohen(1960)}]{cohen1960coefficient}
Jacob Cohen. 1960.
\newblock A coefficient of agreement for nominal scales.
\newblock \emph{Educational and Psychological Measurement}, 20(1):37--46.

\bibitem[{de~Moura and Ullrich(2021)}]{moura2021lean}
Leonardo de~Moura and Sebastian Ullrich. 2021.
\newblock \href {https://doi.org/10.1007/978-3-030-79876-5_37} {The {Lean 4} theorem prover and programming language}.
\newblock In \emph{Automated Deduction -- CADE 28}, volume 12699 of \emph{Lecture Notes in Computer Science}, pages 625--635. Springer.

\bibitem[{Feng et~al.(2026)Feng, Lou, and Sharma}]{feng2026paraphrasefailures}
William Feng, Ethan Lou, and Aryan Sharma. 2026.
\newblock \href {https://arxiv.org/abs/2604.23135v2} {Characterizing paraphrase-induced failures in {Lean 4} autoformalization}.
\newblock \emph{arXiv preprint arXiv:2604.23135}.

\bibitem[{Gao et~al.(2025)Gao, Wang, Jiang, Gao, Qin, Xu, and Dong}]{gao2024herald}
Guoxiong Gao, Yutong Wang, Jiedong Jiang, Qi~Gao, Zihan Qin, Tianyi Xu, and Bin Dong. 2025.
\newblock \href {https://proceedings.iclr.cc/paper_files/paper/2025/hash/8c2bb821410066459be64d03a4dc5719-Abstract-Conference.html} {Herald: A natural language annotated {Lean 4} dataset}.
\newblock In \emph{International Conference on Learning Representations}.

\bibitem[{{Google}(2025)}]{google2025gemini25flash}
{Google}. 2025.
\newblock {Gemini 2.5 Flash}.
\newblock \url{https://cloud.google.com/vertex-ai/generative-ai/docs/models/gemini/2-5-flash}.
\newblock Google Cloud model documentation.

\bibitem[{{Google}(2026)}]{google2026gemini31pro}
{Google}. 2026.
\newblock {Gemini 3.1 Pro}.
\newblock \url{https://docs.cloud.google.com/vertex-ai/generative-ai/docs/models/gemini/3-1-pro}.
\newblock Google Cloud model documentation.

\bibitem[{Guo et~al.(2025)Guo, Yang, Zhang, Song, Wang, Zhu, Xu, Zhang, Ma, Bi et~al.}]{guo2025deepseek}
Daya Guo, Dejian Yang, Haowei Zhang, Junxiao Song, Peiyi Wang, Qihao Zhu, Runxin Xu, Ruoyu Zhang, Shirong Ma, Xiao Bi, et~al. 2025.
\newblock \href {https://doi.org/10.1038/s41586-025-09422-z} {{DeepSeek-R1} incentivizes reasoning in {LLMs} through reinforcement learning}.
\newblock \emph{Nature}, 645(8081):633--638.

\bibitem[{Han et~al.(2022)Han, Rute, Wu, Ayers, and Polu}]{han2021proof}
Jesse~Michael Han, Jason Rute, Yuhuai Wu, Edward~W. Ayers, and Stanislas Polu. 2022.
\newblock \href {https://openreview.net/forum?id=rpXjc9i04U} {Proof artifact co-training for theorem proving with language models}.
\newblock In \emph{The Tenth International Conference on Learning Representations}.

\bibitem[{Hendrycks et~al.(2021)Hendrycks, Burns, Kadavath, Arora, Basart, Tang, Song, and Steinhardt}]{hendrycks2021math}
Dan Hendrycks, Collin Burns, Saurav Kadavath, Akul Arora, Steven Basart, Eric Tang, Dawn Song, and Jacob Steinhardt. 2021.
\newblock Measuring mathematical problem solving with the {MATH} dataset.
\newblock In \emph{Proceedings of the Neural Information Processing Systems Track on Datasets and Benchmarks (NeurIPS Datasets and Benchmarks)}.

\bibitem[{Hu et~al.(2025)Hu, Zhou, Grechuk, and Tyukin}]{hu2025stepproof}
Xiaolin Hu, Qinghua Zhou, Bogdan Grechuk, and Ivan~Y Tyukin. 2025.
\newblock {StepProof}: Step-by-step verification of natural language mathematical proofs.
\newblock \emph{arXiv preprint arXiv:2506.10558}.

\bibitem[{Jana et~al.(2026)Jana, Kale, Tanriverdi, Song, Vishwanath, and Ganesh}]{jana2025proofbridge}
Prithwish Jana, Kaan Kale, Ahmet~Ege Tanriverdi, Cruise Song, Sriram Vishwanath, and Vijay Ganesh. 2026.
\newblock \href {https://proceedings.iclr.cc/paper_files/paper/2026/file/16f5b8d916f13534d11c50adf60914e4-Paper-Conference.pdf} {{ProofBridge}: Auto-formalization of natural language proofs in {Lean} via joint embeddings}.
\newblock In \emph{International Conference on Learning Representations}, pages 13569--13591.

\bibitem[{Jiang et~al.(2024)Jiang, Li, and Jamnik}]{jiang2024multi}
Albert~Q. Jiang, Wenda Li, and Mateja Jamnik. 2024.
\newblock \href {https://doi.org/10.52202/079017-2659} {Multi-language diversity benefits autoformalization}.
\newblock In \emph{Advances in Neural Information Processing Systems}, volume~37, pages 83600--83626.

\bibitem[{Jiang et~al.(2023)Jiang, Welleck, Zhou, Li, Liu, Jamnik, Lacroix, Lample, and Wu}]{jiang2022draft}
Albert~Q. Jiang, Sean Welleck, Jin~Peng Zhou, Wenda Li, Jiacheng Liu, Mateja Jamnik, Timoth{\'{e}}e Lacroix, Guillaume Lample, and Yuhuai Wu. 2023.
\newblock \href {https://openreview.net/forum?id=SMa9EAovKMC} {Draft, sketch, and prove: Guiding formal theorem provers with informal proofs}.
\newblock In \emph{International Conference on Learning Representations}.

\bibitem[{Jiang et~al.(2021)Jiang, Li, Han, and Wu}]{jiang2021lisa}
Albert~Qiaochu Jiang, Wenda Li, Jesse~Michael Han, and Yuhuai Wu. 2021.
\newblock \href {https://aitp-conference.org/2021/abstract/paper_17.pdf} {{LISA}: Language models of {ISAbelle} proofs}.
\newblock In \emph{6th Conference on Artificial Intelligence and Theorem Proving}.
\newblock Conference abstract.

\bibitem[{Kwon et~al.(2023)Kwon, Li, Zhuang, Sheng, Zheng, Yu, Gonzalez, Zhang, and Stoica}]{kwon2023vllm}
Woosuk Kwon, Zhuohan Li, Siyuan Zhuang, Ying Sheng, Lianmin Zheng, Cody~Hao Yu, Joseph~E. Gonzalez, Hao Zhang, and Ion Stoica. 2023.
\newblock Efficient memory management for large language model serving with {PagedAttention}.
\newblock In \emph{Proceedings of the 29th Symposium on Operating Systems Principles}, pages 611--626.

\bibitem[{Lightman et~al.(2024)Lightman, Kosaraju, Burda, Edwards, Baker, Lee, Leike, Schulman, Sutskever, and Cobbe}]{lightman2024let}
Hunter Lightman, Vineet Kosaraju, Yuri Burda, Harrison Edwards, Bowen Baker, Teddy Lee, Jan Leike, John Schulman, Ilya Sutskever, and Karl Cobbe. 2024.
\newblock Let's verify step by step.
\newblock In \emph{International Conference on Learning Representations}, volume 2024, pages 39578--39601.

\bibitem[{Lin et~al.(2026)Lin, Tang, Lyu, Yang, Chung, Zhao, Jiang, Geng, Ge, Sun, Wu, Gesi, Lu, Acuna, Yang, Lin, Choi, Chen, Arora, and Jin}]{lin2025goedelv2}
Yong Lin, Shange Tang, Bohan Lyu, Ziran Yang, Jui-Hui Chung, Haoyu Zhao, Lai Jiang, Yihan Geng, Jiawei Ge, Jingruo Sun, Jiayun Wu, Jiri Gesi, Ximing Lu, David Acuna, Kaiyu Yang, Hongzhou Lin, Yejin Choi, Danqi Chen, Sanjeev Arora, and Chi Jin. 2026.
\newblock \href {https://proceedings.iclr.cc/paper_files/paper/2026/file/13e8be77982beb73d7ed0bbf122f9f3c-Paper-Conference.pdf} {{Goedel-Prover-V2}: Scaling formal theorem proving with scaffolded data synthesis and self-correction}.
\newblock In \emph{International Conference on Learning Representations}, pages 11793--11818.

\bibitem[{Liu et~al.(2025{\natexlab{a}})Liu, Yuan, Yin, Xu, Xu, Chen, Wang, Shang, Liu, and Zhang}]{liu2025safe}
Chengwu Liu, Ye~Yuan, Yichun Yin, Yan Xu, Xin Xu, Zaoyu Chen, Yasheng Wang, Lifeng Shang, Qun Liu, and Ming Zhang. 2025{\natexlab{a}}.
\newblock \href {https://doi.org/10.18653/v1/2025.acl-long.594} {{SAFE}: Enhancing mathematical reasoning in large language models via retrospective step-aware formal verification}.
\newblock In \emph{Proceedings of the 63rd Annual Meeting of the Association for Computational Linguistics (Volume 1: Long Papers)}, pages 12171--12186. Association for Computational Linguistics.

\bibitem[{Liu et~al.(2025{\natexlab{b}})Liu, Zheng, Lu, Cao, and Yan}]{liu2025rethinking}
Qi~Liu, Xinhao Zheng, Xudong Lu, Qinxiang Cao, and Junchi Yan. 2025{\natexlab{b}}.
\newblock Rethinking and improving autoformalization: Towards a faithful metric and a dependency retrieval-based approach.
\newblock In \emph{The Thirteenth International Conference on Learning Representations}.

\bibitem[{Liu and Zou(2026)}]{liu2026explanations}
Yang Liu and Ran Zou. 2026.
\newblock \href {https://arxiv.org/abs/2608.12623} {When explanations betray backdoors: Black-box auditing for language model classifiers}.
\newblock \emph{arXiv preprint arXiv:2608.12623}.

\bibitem[{Lu et~al.(2024)Lu, Wan, Liu, Huang, Xiong, Liu, Shen, Jin, Zhang, Wang et~al.}]{lu2024process}
Jianqiao Lu, Yingjia Wan, Zhengying Liu, Yinya Huang, Jing Xiong, Chengwu Liu, Jianhao Shen, Hui Jin, Jipeng Zhang, Haiming Wang, et~al. 2024.
\newblock \href {https://arxiv.org/abs/2406.01940v1} {Process-driven autoformalization in {Lean 4}}.
\newblock \emph{arXiv preprint arXiv:2406.01940}.
\newblock Version 1, June 4, 2024.

\bibitem[{Menick et~al.(2022)Menick, Trebacz, Mikulik, Aslanides, Song, Chadwick, Glaese, Young, Campbell-Gillingham, Irving, and McAleese}]{menick2022teaching}
Jacob Menick, Maja Trebacz, Vladimir Mikulik, John Aslanides, Francis Song, Martin Chadwick, Mia Glaese, Susannah Young, Lucy Campbell-Gillingham, Geoffrey Irving, and Nat McAleese. 2022.
\newblock Teaching language models to support answers with verified quotes.
\newblock \emph{arXiv preprint arXiv:2203.11147}.

\bibitem[{Moore and Shah(2025)}]{moore2025evaluating}
Hayden Moore and Asfahan Shah. 2025.
\newblock Evaluating autoformalization robustness via semantically similar paraphrasing.
\newblock \emph{arXiv preprint arXiv:2511.12784}.

\bibitem[{Nipkow et~al.(2002)Nipkow, Paulson, and Wenzel}]{nipkow2002isabelle}
Tobias Nipkow, Lawrence~C. Paulson, and Markus Wenzel. 2002.
\newblock \href {https://doi.org/10.1007/3-540-45949-9} {\emph{{Isabelle/HOL}: A Proof Assistant for Higher-Order Logic}}, volume 2283 of \emph{Lecture Notes in Computer Science}.
\newblock Springer.

\bibitem[{{OpenAI}(2026)}]{openai2026gpt54mini}
{OpenAI}. 2026.
\newblock {GPT-5.4} mini.
\newblock \url{https://developers.openai.com/api/docs/models/gpt-5.4-mini}.
\newblock OpenAI API model documentation.

\bibitem[{Poiroux et~al.(2025)Poiroux, Weiss, Kun{\v{c}}ak, and Bosselut}]{poiroux2025reliable}
Auguste Poiroux, Gail Weiss, Viktor Kun{\v{c}}ak, and Antoine Bosselut. 2025.
\newblock \href {https://doi.org/10.18653/v1/2025.emnlp-main.907} {Reliable evaluation and benchmarks for statement autoformalization}.
\newblock In \emph{Proceedings of the 2025 Conference on Empirical Methods in Natural Language Processing}, pages 17947--17969. Association for Computational Linguistics.

\bibitem[{Polu and Sutskever(2020)}]{polu2020generative}
Stanislas Polu and Ilya Sutskever. 2020.
\newblock Generative language modeling for automated theorem proving.
\newblock \emph{arXiv preprint arXiv:2009.03393}.

\bibitem[{{Qwen Team}(2026)}]{qwen35ModelCard}
{Qwen Team}. 2026.
\newblock {Qwen3.5-397B-A17B}.
\newblock \url{https://huggingface.co/Qwen/Qwen3.5-397B-A17B}.
\newblock Hugging Face model card.

\bibitem[{Ren et~al.(2025)Ren, Shao, Song, Xin, Wang, Zhao, Zhang, Fu, Zhu, Yang et~al.}]{ren2025deepseek}
ZZ~Ren, Zhihong Shao, Junxiao Song, Huajian Xin, Haocheng Wang, Wanjia Zhao, Liyue Zhang, Zhe Fu, Qihao Zhu, Dejian Yang, et~al. 2025.
\newblock {DeepSeek-Prover-V2}: Advancing formal mathematical reasoning via reinforcement learning for subgoal decomposition.
\newblock \emph{arXiv preprint arXiv:2504.21801}.

\bibitem[{{The mathlib Community}(2020)}]{mathlib2020}
{The mathlib Community}. 2020.
\newblock The {Lean} mathematical library.
\newblock In \emph{Proceedings of the 9th ACM SIGPLAN International Conference on Certified Programs and Proofs}, pages 367--381.

\bibitem[{Tsoukalas et~al.(2024)Tsoukalas, Lee, Jennings, Xin, Ding, Jennings, Thakur, and Chaudhuri}]{tsoukalas2024putnambench}
George Tsoukalas, Jasper Lee, John Jennings, Jimmy Xin, Michelle Ding, Michael Jennings, Amitayush Thakur, and Swarat Chaudhuri. 2024.
\newblock {PutnamBench}: Evaluating neural theorem-provers on the {Putnam} mathematical competition.
\newblock \emph{Advances in Neural Information Processing Systems}, 37:11545--11569.

\bibitem[{Wang et~al.(2025)Wang, Unsal, Lin, Baksys, Liu, Santos, Sung, Vinyes, Ying, Zhu et~al.}]{wang2025kimina}
Haiming Wang, Mert Unsal, Xiaohan Lin, Mantas Baksys, Junqi Liu, Marco~Dos Santos, Flood Sung, Marina Vinyes, Zhenzhe Ying, Zekai Zhu, et~al. 2025.
\newblock {Kimina-Prover Preview}: Towards large formal reasoning models with reinforcement learning.
\newblock \emph{arXiv preprint arXiv:2504.11354}.

\bibitem[{Wang et~al.(2024)Wang, Zhang, Jia, Pan, Diao, Pi, and Zhang}]{wang2024theoremllama}
Ruida Wang, Jipeng Zhang, Yizhen Jia, Rui Pan, Shizhe Diao, Renjie Pi, and Tong Zhang. 2024.
\newblock {TheoremLlama}: Transforming general-purpose {LLMs} into {Lean4} experts.
\newblock In \emph{Proceedings of the 2024 Conference on Empirical Methods in Natural Language Processing}, pages 11953--11974.

\bibitem[{Wilson(1927)}]{wilson1927probable}
Edwin~B. Wilson. 1927.
\newblock Probable inference, the law of succession, and statistical inference.
\newblock \emph{Journal of the American Statistical Association}, 22(158):209--212.

\bibitem[{Wu et~al.(2022)Wu, Jiang, Li, Rabe, Staats, Jamnik, and Szegedy}]{wu2022autoformalization}
Yuhuai Wu, Albert~Qiaochu Jiang, Wenda Li, Markus Rabe, Charles Staats, Mateja Jamnik, and Christian Szegedy. 2022.
\newblock Autoformalization with large language models.
\newblock \emph{Advances in Neural Information Processing Systems}, 35:32353--32368.

\bibitem[{Wu et~al.(2026)Wu, Huang, Wan, Peng, Shang, Cao, Qi, Zhang, Zhang, Du et~al.}]{wu2026stepfun}
Yutong Wu, Di~Huang, Ruosi Wan, Yue Peng, Shijie Shang, Chenrui Cao, Lei Qi, Rui Zhang, Xishan Zhang, Zidong Du, et~al. 2026.
\newblock {StepFun-Formalizer}: Unlocking the autoformalization potential of {LLMs} through knowledge-reasoning fusion.
\newblock In \emph{Proceedings of the AAAI Conference on Artificial Intelligence}, volume~40, pages 33980--33988.

\bibitem[{Xin et~al.(2024)Xin, Guo, Shao, Ren, Zhu, Liu, Ruan, Li, and Liang}]{xin2024deepseek}
Huajian Xin, Daya Guo, Zhihong Shao, Zhizhou Ren, Qihao Zhu, Bo~Liu, Chong Ruan, Wenda Li, and Xiaodan Liang. 2024.
\newblock {DeepSeek-Prover}: Advancing theorem proving in {LLMs} through large-scale synthetic data.
\newblock \emph{arXiv preprint arXiv:2405.14333}.

\bibitem[{Yao et~al.(2025)Yao, Wang, and Zhang}]{yao-etal-2025-fans}
Jiarui Yao, Ruida Wang, and Tong Zhang. 2025.
\newblock \href {https://doi.org/10.18653/v1/2025.emnlp-main.158} {{FANS}: Formal answer selection for {LLM} natural language math reasoning using {Lean4}}.
\newblock In \emph{Proceedings of the 2025 Conference on Empirical Methods in Natural Language Processing}, pages 3181--3200. Association for Computational Linguistics.

\bibitem[{Ying et~al.(2024)Ying, Wu, Geng, Wang, Lin, and Chen}]{ying2024lean}
Huaiyuan Ying, Zijian Wu, Yihan Geng, Jiayu Wang, Dahua Lin, and Kai Chen. 2024.
\newblock {Lean Workbook}: A large-scale {Lean} problem set formalized from natural language math problems.
\newblock \emph{Advances in Neural Information Processing Systems}, 37:105848--105863.

\bibitem[{Zheng et~al.(2022)Zheng, Han, and Polu}]{zheng2021minif2f}
Kunhao Zheng, Jesse~Michael Han, and Stanislas Polu. 2022.
\newblock \href {https://openreview.net/forum?id=9ZPegFuFTFv} {{miniF2F}: A cross-system benchmark for formal olympiad-level mathematics}.
\newblock In \emph{International Conference on Learning Representations}.

\end{thebibliography}

\clearpage
\appendix
\crefalias{section}{appendix}\crefalias{subsection}{appendix}\crefalias{subsubsection}{appendix}
\raggedbottom
\tcbset{breakable}
\setcounter{dbltopnumber}{2}
\renewcommand{\dbltopfraction}{0.95}
\renewcommand{\textfraction}{0.05}
\renewcommand{\dblfloatpagefraction}{0.8}
\makeatletter
\setlength{\@fptop}{0pt}
\setlength{\@dblfptop}{0pt}
\setlength{\@fpsep}{12pt plus 2pt minus 2pt}
\setlength{\@dblfpsep}{12pt plus 2pt minus 2pt}
\makeatother

\section{Implementation Details}
\label{app:implementation-detail}

\subsection{Construction of Perturbations}
\label{app:perturbation-construction}

\paragraph{Global perturbations.}
We obtain global perturbations by querying Gemini-2.5-Flash and Qwen3.5-397B-A17B with the free-form and step-by-step paraphrasing prompts in \Cref{app:rephrase-prompt}.

\paragraph{Local perturbations.}
All three local edit types share the same three-stage pipeline. First, an LLM (Gemini-2.5-Flash) marks candidate edit locations. Second, we filter the candidates and pick one per region, where a region is either the statement or the proof. Third, a deterministic rule applies the edit as a single-span change at the recorded location.

\paragraph{Number edit.}
An LLM marks each editable number in the statement and proof, with structural numbers (subscripts, superscripts, indices) and numbers inside LaTeX formatting arguments skipped. A SHA256-seeded randomization then sets the new value: it shifts the number by an amount that scales with its size (for example, an integer up to $10$ changes by $1$ to $3$, and one up to $100$ by $1$ to $5$) and keeps the sign, so that the edited problem stays well-formed.

\paragraph{Symbol edit.}
An LLM marks relation and operator symbols (other than $=$ and $\neq$) and tags each as a relation or an operator. We swap each one using a fixed table: relations flip direction ($<\!\leftrightarrow\!>$, $\leq\!\leftrightarrow\!\geq$) and operators go to their inverse ($+\!\leftrightarrow\!-$, $\times\!\leftrightarrow\!\div$). We skip symbols inside relation chains (e.g., $0<p<15$) and inside subscripts or superscripts, and we oversample relations to balance relation and operator symbols.

\paragraph{Step deletion.}
\label{app:deletion-interpretation}
An LLM first splits each proof into steps, each step with a reasoning part and an outcome, and flags a step as deletable only if two conditions hold. First, its reasoning is substantive: it applies a theorem or principle, carries out a calculation or algebraic transformation, or derives an intermediate fact, rather than serving as a transition such as ``thus'' or ``we have''. Second, its outcome stands on its own once the reasoning is removed: it is a contiguous part of the step that reads as a complete claim, not a fragment that depends on the deleted text (such as one beginning with ``which'' or ``this gives'') and not a bare number or symbol. Reasoning shorter than 30 characters is kept only if it contains a mathematical expression or an explicit reasoning action.
We then select one deletable step per proof and delete its reasoning while retaining its outcome.
When judging, an output that supplies tactics deriving the retained outcome is labeled reverted, even if the derivation is a single short tactic and regardless of whether it type-checks; the full criteria are in the judge prompt in \Cref{app:edit-judge-prompts}.

\begin{table}[t]
\centering

\adjustbox{max width=.40\textwidth}{
\begin{tabular}{lcc}
\toprule
& \textbf{miniF2F} & \textbf{MATH-500} \\
\midrule
Number edit   & 227 / 241 & 490 / 489 \\
Symbol edit   & 130 / 195 & 195 / 349 \\
Step deletion & --\phantom{00} / 129 & --\phantom{00} / 192 \\
\bottomrule
\end{tabular}}
\caption{Dataset statistics for local perturbations, reporting the number of statement-level / proof-level edit instances on miniF2F and MATH-500.}
\label{tab:edit-counts}
\end{table}

\subsection{Compute}
\label{app:experimental-setup}

We use vLLM~\citep{kwon2023vllm} to serve open-source models and query APIs for commercial ones. All models other than ProofFlow generate with temperature $0.6$ and top-$p$ $0.95$.
All main experiments run on a cluster with NVIDIA A40, A100, or H200 GPUs. Runtime does not affect our results, so we use whichever of these GPUs is available. We estimate the project's total cost at about 1,900 GPU-hours.
For type checking, we use Lean~4.15.0 and Mathlib~\citep{mathlib2020} v4.15.0, through a persistent Lean REPL that keeps Mathlib loaded across calls. Each type-check has a per-problem timeout of $900$\,seconds.

For ProofFlow, we use the released dependency-DAG pipeline and adapt its component models and output interface to our evaluation: the graph builder is Gemini-2.5-Flash, and the lemma formalizer and tactic completer are \texttt{Goedel-LM/Goedel-Formalizer-V2-8B} and \texttt{Goedel-LM/Goedel-Prover-V2-8B}~\citep{lin2025goedelv2}. We use three retries for graph construction, lemma formalization, and proving, and all its components generate with temperature $0.9$ and top-$p$ $0.95$.

\subsection{Licenses and Terms}
The Lean~4 toolchain~\citep{moura2021lean}, Mathlib, miniF2F~\citep{zheng2021minif2f}, and all open-weight models we use are released under Apache-2.0, except that the DeepSeek-Prover-V2-7B weights~\citep{ren2025deepseek} are under the DeepSeek Model License. MATH~\citep{hendrycks2021math} and the ProofBridge~\citep{jana2025proofbridge} and ProofFlow~\citep{cabral2025proofflow} codebases are released under the MIT license. We access the Gemini models through the Gemini API, and through OpenRouter for the step-deletion judge, under their terms. Our usage is consistent with their licenses and terms, and we release our code and data under the MIT license.

\section{Prompt Templates}

This section collects all prompts we use: the paraphrasing prompts for global perturbations, the autoformalization prompts, and the LLM judge prompts for semantic correctness and faithfulness.

\subsection{Prompts for Global Perturbations}

We first give the rules shared by both paraphrasing modes, and then the prompts for free-form and step-by-step paraphrasing.

\paragraph{Shared paraphrasing rules.} These rules are inserted into both paraphrasing prompts below, in place of \texttt{\{faithfulness\_rules\}}. They require the paraphrase to keep the original meaning, length, and logical steps, and to avoid fixing or extending the original.
\begin{tcolorbox}[
    colback=gray!5,
    colframe=gray!60,
    coltitle=white,
    colbacktitle=gray!65,
    title=\textbf{Shared Paraphrasing Rules},
    fonttitle=\bfseries,
    arc=2mm,
    boxrule=0.6pt,
    left=2mm,
    right=2mm,
    top=1mm,
    bottom=1mm
]
\ttfamily\small
This is a MEANING-PRESERVING transformation. Your rewrite must be a faithful translation of the original --- a reader must be able to recover the original's exact mathematical content from your version. Specifically:

\medskip

- Do NOT change any numerical constant, variable name, function name, set, inequality direction, or quantifier.

- Do NOT change, weaken, strengthen, or generalize the hypotheses or the conclusion.

- Do NOT add new mathematical facts, lemmas, intermediate results, or justifications that were not in the original.

- Do NOT remove, merge, or reorder logical steps of the original argument.

- Do NOT fix or ``clean up'' the original. If the original is awkward, incomplete, or even wrong, preserve it faithfully --- your job is translation, not editing.

- The rewritten version should be roughly the same length as the original, within \(\pm 30\%\).

\medskip

Use LaTeX math notation exactly as in the original, e.g., \(x^2\), \textbackslash frac\{a\}\{b\}, \textbackslash neq, \textbackslash notin, \textbackslash tan. Write every backslash literally as a single backslash --- do NOT escape them, do NOT wrap anything in JSON, do NOT add code fences.

\medskip

Output format --- return EXACTLY these two tagged blocks and NOTHING else before or after. Your entire response must start with <informal\_statement> and end with </informal\_proof>.

\medskip

<informal\_statement>

... rewritten problem statement ...

</informal\_statement>

\medskip

<informal\_proof>

... rewritten proof ...

</informal\_proof>
\end{tcolorbox}

\label{app:rephrase-prompt}

We use the following prompt for free-form paraphrasing (G-FF and Q-FF).

\begin{tcolorbox}[
    colback=gray!5,
    colframe=gray!60,
    coltitle=white,
    colbacktitle=gray!65,
    title=\textbf{Free-Form Paraphrasing Prompt},
    fonttitle=\bfseries,
    arc=2mm,
    boxrule=0.6pt,
    left=2mm,
    right=2mm,
    top=1mm,
    bottom=1mm
]
\ttfamily\small
You are a mathematics professor. Paraphrase the following problem statement and its proof into different wording while preserving the exact same mathematical meaning. Change only the phrasing, sentence structure, and word choice; keep every mathematical object, value, and logical step identical.

\medskip

\{faithfulness\_rules\}

\medskip

Original problem statement:

\%STATEMENT\%

\medskip

Original proof:

\%PROOF\%
\end{tcolorbox}

\label{app:step-prompt}
We use the following prompt for step-by-step paraphrasing (G-Step and Q-Step), which reorganizes the NL theorem--proof pair into a numbered sequence of steps.

\begin{tcolorbox}[
    colback=gray!5,
    colframe=gray!60,
    coltitle=white,
    colbacktitle=gray!65,
    title=\textbf{Step-by-Step Paraphrasing Prompt},
    fonttitle=\bfseries,
    arc=2mm,
    boxrule=0.6pt,
    left=2mm,
    right=2mm,
    top=1mm,
    bottom=1mm
]
\ttfamily\scriptsize
You are a mathematics professor. Faithfully translate the following problem statement and its proof into a numbered step-by-step format. This is a pure formatting change: break the existing content into numbered steps with short bold step titles. Do NOT add explanations, details, motivation, or any content that is not already in the original --- only reorganize what is there.

\medskip

The proof should follow this structure:

1. **Step Title:** <content taken directly from the original>

2. **Step Title:** <content taken directly from the original>

\medskip

The statement may use the same numbered-step format if it naturally decomposes, or remain as a single paragraph if it does not --- but its mathematical content must stay identical.

\medskip

\{faithfulness\_rules\}

\medskip

Original problem statement:

\%STATEMENT\%

\medskip

Original proof:

\%PROOF\%
\end{tcolorbox}

\subsection{Autoformalization Prompts}

We use the following system and user prompts to autoformalize each NL theorem--proof pair into an FL output, i.e., a Lean~4 theorem statement together with its proof.

\begin{tcolorbox}[
    colback=gray!5,
    colframe=gray!60,
    coltitle=white,
    colbacktitle=gray!65,
    title=\textbf{Autoformalization System Prompt},
    fonttitle=\bfseries,
    arc=2mm,
    boxrule=0.6pt,
    left=2mm,
    right=2mm,
    top=1mm,
    bottom=1mm
]
\ttfamily\small
You are an expert in mathematics. Your task is to convert informal, natural-language proofs into correct Lean 4 formalizations.
\end{tcolorbox}

\begin{tcolorbox}[
    colback=gray!5,
    colframe=gray!60,
    coltitle=white,
    colbacktitle=gray!65,
    title=\textbf{Autoformalization User Prompt},
    fonttitle=\bfseries,
    arc=2mm,
    boxrule=0.6pt,
    left=2mm,
    right=2mm,
    top=1mm,
    bottom=1mm
]
\ttfamily\small
Your task is to take as input an informal proof in natural language and autoformalize it in Lean 4 with a header.
Think step-by-step and ensure that the output formal theorem is compilable with Lean 4 (version 4.15.0).

\medskip

Here is the **actual** informal proof in natural language:

\medskip

<informal\_statement>

\{informal\_statement\}

</informal\_statement>

\medskip

<informal\_proof>

\{informal\_proof\}

</informal\_proof>

\medskip

Now first think step-by-step for the actual output and autoformalize it in Lean 4 with a header. Importantly, enclose the final formal proof in Lean 4 inside the following tags:

\medskip

<formal\_proof>

\textasciigrave\textasciigrave\textasciigrave lean4

(Provide your entire Lean 4 proof with header here)

\textasciigrave\textasciigrave\textasciigrave

</formal\_proof>
\end{tcolorbox}

\subsection{Semantic Correctness Judge Prompts}
\label{app:sc-judge-prompts}

We use two judge prompts, which check whether the FL proof and the FL statement are semantically consistent with the NL proof and the NL theorem, respectively (ProofSC and StmtSC).

\begin{tcolorbox}[
    colback=gray!5,
    colframe=gray!60,
    coltitle=white,
    colbacktitle=gray!65,
    title=\textbf{Proof Semantic Correctness Judge Prompt},
    fonttitle=\bfseries,
    arc=2mm,
    boxrule=0.6pt,
    left=2mm,
    right=2mm,
    top=1mm,
    bottom=1mm
]
\ttfamily\small
You are an expert in formal mathematics and Lean 4. Judge ONE thing:

Do the following two pieces of reasoning carry out the SAME LOGICAL STEPS?

\medskip

=== CRITICAL --- Independence from validity ===

\medskip

You are shown a natural-language proof and a sequence of Lean 4 proof tactics. You are NOT shown the Lean theorem statement that these tactics target. This is intentional and important.

Your job is to compare the two REASONING SEQUENCES on their own merit:

\medskip

- Whether the Lean tactics actually prove the hidden Lean theorem is checked separately by Lean's type checker. You MUST IGNORE that question entirely. The tactics may or may not prove the theorem they are attached to --- that is not your concern.

\medskip

- Even if the Lean tactics seem to refer to a different goal than the NL proof suggests, evaluate the reasoning sequence on its own terms.

\medskip

=== Equivalence criteria ===

\medskip

Two proofs are equivalent if they follow the SAME OVERALL STRATEGY and the substantive intermediate claims agree, modulo:

- wording and step ordering,

- level of abstraction. Lean tactics may IMPLEMENT abstract NL principles such as AM-GM, Cauchy-Schwarz, or trigonometric identities via automation tactics like \texttt{nlinarith}, \texttt{polyrith}, or \texttt{positivity}. This counts as the SAME step, not a different one.

\medskip

They are NOT equivalent if:

- they use a fundamentally different strategy or prove a different goal,

- the Lean proof body is empty, \texttt{sorry}, \texttt{admit}, or a degenerate placeholder,

- the substantive intermediate claims disagree.

\medskip

=== Output ===

\medskip

First, briefly map each substantive NL step to a corresponding Lean tactic, or note where they disagree. Then output:

\medskip

<proof\_justification>your step-by-step correspondence here</proof\_justification>

<proof\_score>1 or -1</proof\_score>

\medskip

=== Inputs ===

\medskip

Natural-language proof:

\{nl\_proof\}

\medskip

Lean 4 proof tactics, with theorem signature deliberately hidden:

\{lean\_proof\_body\}
\end{tcolorbox}

\begin{tcolorbox}[
    colback=gray!5,
    colframe=gray!60,
    coltitle=white,
    colbacktitle=gray!65,
    title=\textbf{Statement Semantic Correctness Judge Prompt},
    fonttitle=\bfseries,
    arc=2mm,
    boxrule=0.6pt,
    left=2mm,
    right=2mm,
    top=1mm,
    bottom=1mm
]
\ttfamily\small
You are an expert in formal mathematics and Lean 4.

Does the following Lean 4 theorem signature express the same mathematical proposition as the natural-language statement?

\medskip

=== Equivalence criteria ===

\medskip

Two statements are equivalent if they capture the same mathematical content, even if they differ in surface form: variable names, quantifier order when commutative, notation such as \texttt{\textbackslash le} vs \(\leq\), \texttt{\textbackslash sum} vs \(\sum\), or how the conclusion is phrased.

Use your mathematical understanding to judge whether the two statements describe the same problem.

\medskip

=== Note on inputs ===

\medskip

You are shown the Lean theorem signature, including binders and goal type, only; the Lean proof body is hidden because it is judged separately.

\medskip

=== Output ===

\medskip

Provide a clear justification, then output:

\medskip

<stmt\_justification>your reasoning here</stmt\_justification>

<stmt\_score>1 or -1</stmt\_score>

\medskip

=== Inputs ===

\medskip

Natural-language statement:

\{nl\_statement\}

\medskip

Lean 4 theorem signature:

\{lean\_theorem\_signature\}
\end{tcolorbox}

\subsection{Faithfulness Judge Prompts}
\label{app:edit-judge-prompts}
We use one judge prompt for each local perturbation type. Each prompt labels the FL output as ``faithful'', ``reverted'', or ``other'', where ``other'' corresponds to unclear outputs (\Cref{sec:faithfulness-metrics}).

\paragraph{Number edit.} This prompt decides whether the FL output reflects the edited number.
\begin{tcolorbox}[
    colback=gray!5,
    colframe=gray!60,
    coltitle=white,
    colbacktitle=gray!65,
    title=\textbf{Number-Edit Faithfulness Judge Prompt},
    fonttitle=\bfseries,
    arc=2mm,
    boxrule=0.6pt,
    left=2mm,
    right=2mm,
    top=1mm,
    bottom=1mm
]
\ttfamily\small
You are an expert in mathematical formalization. Given a math problem (whose natural language has been edited by changing one number) and a fragment of the model-generated Lean 4 formalization, determine whether the model faithfully reflects the number edit.

\medskip

The Lean fragment shown below is the ONLY part of the formalization you should consider. It is the region targeted by the edit (\{source\}). Do not infer or imagine any other Lean code. Base your judgment strictly on the fragment.

\medskip

=== Edit Info ===

\medskip

Edit type: \{edit\_type\}; Original number: \{old\_value\}; Edited number: \{new\_value\}; Edit location: \{source\}; Edit context: ``\{context\}''.

\medskip

=== Inputs ===

\medskip

Edited informal statement: \{informal\_statement\}\\
Edited informal proof: \{informal\_proof\}\\
Model-generated Lean 4 fragment (\{source\} region only): \{generated\_fl\}

\medskip

=== Tasks ===

\medskip

1. In the Lean fragment, find the number that semantically corresponds to ``\{new\_value\}'' or ``\{old\_value\}'' in the edit context. Numbers in Lean may differ from natural language; e.g., \(6.5\) may appear as \(13/2\), and \(65\) may appear as \texttt{(65 : R)}.

\medskip

2. Determine the number at that position: if it is \{new\_value\} (or an equivalent form) \(\rightarrow\) ``faithful''; if it is \{old\_value\} (or an equivalent form) \(\rightarrow\) ``reverted''; if the corresponding position cannot be found, or it is some other value \(\rightarrow\) ``other''.

\medskip

\textbf{IMPORTANT} --- ground your judgment in evidence; do not hallucinate. Before returning ``faithful'' or ``reverted'', you MUST be able to point to the literal substring of the fragment where the value appears. If you cannot find such a substring, return \texttt{value\_in\_fl=""} and \texttt{judgment="other"}. The \texttt{context\_in\_fl} field below is the proof --- it must be copied verbatim from the fragment and must contain \texttt{value\_in\_fl}.

\medskip

=== Output (strict JSON only) ===

\medskip

\{``found\_in\_fl'': true/false,\\
\hspace*{1em}``value\_in\_fl'': ``copy the number from the fragment as-is (or empty string if not found)'',\\
\hspace*{1em}``context\_in\_fl'': ``if value\_in\_fl is non-empty, copy the \(\sim\)40-character window surrounding the value; otherwise empty string'',\\
\hspace*{1em}``judgment'': ``faithful''/``reverted''/``other'',\\
\hspace*{1em}``reason'': ``one-sentence explanation''\}
\end{tcolorbox}

\paragraph{Symbol edit.} This prompt decides whether the FL output reflects the edited symbol.
\begin{tcolorbox}[
    colback=gray!5,
    colframe=gray!60,
    coltitle=white,
    colbacktitle=gray!65,
    title=\textbf{Symbol-Edit Faithfulness Judge Prompt},
    fonttitle=\bfseries,
    arc=2mm,
    boxrule=0.6pt,
    left=2mm,
    right=2mm,
    top=1mm,
    bottom=1mm
]
\ttfamily\small
You are an expert in mathematical formalization. Given a math problem (whose natural language has been edited by changing one math symbol) and a fragment of the model-generated Lean 4 formalization, determine whether the model faithfully reflects the symbol edit.

\medskip

The Lean fragment shown below is the ONLY part of the formalization you should consider. It is the region targeted by the edit (\{source\}). Do not infer or imagine any other Lean code. Base your judgment strictly on the fragment.

\medskip

=== Edit Info ===

\medskip

Edit type: \{edit\_type\}; Original symbol: \{old\_symbol\}; Edited symbol: \{new\_symbol\}; Symbol family: \{family\}; Edit location: \{source\}; Edit context: ``\{context\}''.

\medskip

=== Inputs ===

\medskip

Edited informal statement: \{informal\_statement\}\\
Edited informal proof: \{informal\_proof\}\\
Model-generated Lean 4 fragment (\{source\} region only): \{generated\_fl\}

\medskip

=== Tasks ===

\medskip

1. In the Lean fragment, find the math symbol corresponding to the edited symbol in the edit context. Symbol mapping from LaTeX to Lean: \texttt{\textbackslash geq} or \texttt{>=} \(\rightarrow\) \texttt{>=}; \texttt{\textbackslash leq} or \texttt{<=} \(\rightarrow\) \texttt{<=}; \texttt{>} \(\rightarrow\) \texttt{>}; \texttt{<} \(\rightarrow\) \texttt{<}; \texttt{\textbackslash neq} or \texttt{!=} \(\rightarrow\) \texttt{!=} or \texttt{Ne}; \texttt{+} \(\rightarrow\) \texttt{+}; \texttt{-} \(\rightarrow\) \texttt{-}; \texttt{\textbackslash times} or \texttt{\textbackslash cdot} \(\rightarrow\) \texttt{*}; \texttt{\textbackslash div} \(\rightarrow\) \texttt{/}. Note: \texttt{a > b} in Lean may equivalently be written as \texttt{b < a}.

\medskip

2. Determine the symbol at that position: if it is the Lean equivalent of \{new\_symbol\} \(\rightarrow\) ``faithful''; if it is the Lean equivalent of \{old\_symbol\} \(\rightarrow\) ``reverted''; otherwise \(\rightarrow\) ``other''.

\medskip

\textbf{IMPORTANT} --- ground your judgment in evidence; do not hallucinate. Before returning ``faithful'' or ``reverted'', you MUST be able to point to the literal substring of the fragment where the symbol appears. If you cannot find such a substring, return \texttt{value\_in\_fl=""} and \texttt{judgment="other"}. The \texttt{context\_in\_fl} field below is the proof --- it must be copied verbatim from the fragment.

\medskip

=== Output (strict JSON only) ===

\medskip

\{``found\_in\_fl'': true/false,\\
\hspace*{1em}``value\_in\_fl'': ``copy the symbol from the fragment as-is (or empty string if not found)'',\\
\hspace*{1em}``context\_in\_fl'': ``\(\sim\)40-character window surrounding the symbol, or empty string'',\\
\hspace*{1em}``judgment'': ``faithful''/``reverted''/``other'',\\
\hspace*{1em}``reason'': ``one-sentence explanation''\}
\end{tcolorbox}

\paragraph{Step deletion.} This prompt decides whether the FL output retains the gap left by the deleted step or instead supplies a justification for the retained claim. We run it with Gemini-2.5-Flash at temperature $0$.
\begin{tcolorbox}[
    colback=gray!5,
    colframe=gray!60,
    coltitle=white,
    colbacktitle=gray!65,
    title=\textbf{Step-Deletion Faithfulness Judge Prompt},
    fonttitle=\bfseries,
    arc=2mm,
    boxrule=0.6pt,
    left=2mm,
    right=2mm,
    top=1mm,
    bottom=1mm
]
\ttfamily\small
You are an expert in mathematical formalization and Lean 4. A natural-language proof was edited by deleting the justification of one proof step while keeping that step's stated outcome (the ``retained claim''). A model then produced a Lean 4 formalization from the edited proof.

\medskip

Decide whether the Lean code RETAINS the gap left by the deletion or SUPPLIES a justification for the retained claim.

\medskip

=== Edit Info ===

\medskip

Deleted justification: ``\{reasoning\_text\}''; Retained claim: ``\{outcome\_text\}''; The edited step was the last step of the proof: \{is\_last\}.

\medskip

=== Inputs ===

\medskip

Edited informal proof (justification removed): \{informal\_proof\}\\
Model-generated Lean 4 code: \{generated\_fl\}

\medskip

=== Rules ===

\medskip

Comments are not code. Text inside \texttt{-{}- ...} line comments or \texttt{/- ... -/} block comments must be ignored when locating the claim and must never be used as evidence.

\medskip

1. Locate where the retained claim is formalized in the Lean code: a \texttt{have}, \texttt{show}, \texttt{obtain}, \texttt{calc} step, \texttt{suffices}, the theorem conclusion, or an equivalent statement. If the last step was edited, the retained claim may be the theorem's conclusion.

\medskip

2. Classify how the code handles that claim. Judge what the code supplies, not whether it would type-check: compilation is checked separately by Lean, so do not try to decide whether a tactic would succeed. ``unproved'': the claim is stated but given no proof, i.e.\ an explicit placeholder such as \texttt{:= by sorry}, \texttt{:= sorry}, \texttt{admit}, or \texttt{?\_}. Never choose ``unproved'' because you think a supplied tactic is insufficient, the claim is false, or the code would not compile; if tactics are written for the claim, it is ``derived''. ``assumed'': the claim is taken as given rather than derived, e.g.\ it appears as a hypothesis or axiom, or it is used later without ever being proved. ``derived'': the code supplies a proof for the claim. Any tactic proof written for the claim counts, regardless of length and regardless of whether it would compile. A single automation tactic such as \texttt{nlinarith [sq\_nonneg (x - 3)]}, \texttt{linarith}, \texttt{norm\_num}, \texttt{simp}, \texttt{ring\_nf}, \texttt{field\_simp}, \texttt{omega}, \texttt{decide}, \texttt{positivity}, or \texttt{exact <lemma>} supplies the justification when it closes the claim. Do not treat a proof as missing merely because it is short. ``absent'': the claim cannot be located in the code, or the output is empty or an error.

\medskip

3. Map to a judgment: ``unproved'' or ``assumed'' \(\rightarrow\) ``faithful'' (the gap is retained); ``derived'' \(\rightarrow\) ``reverted'' (the justification is supplied); ``absent'' \(\rightarrow\) ``other''.

\medskip

\textbf{IMPORTANT}: ground your judgment in evidence. For ``faithful'' or ``reverted'', copy verbatim a snippet of Lean CODE (not a comment) of at most about 80 characters that shows how the claim is handled: the \texttt{sorry}/\texttt{?\_}/hypothesis for ``faithful'', or the claim together with the tactic that proves it for ``reverted''. If you cannot point to such code, return \texttt{handling="absent"} and \texttt{judgment="other"}.

\medskip

=== Output (strict JSON only) ===

\medskip

\{``claim\_location'': ``copy the Lean code line where the retained claim is stated, or empty string'',\\
\hspace*{1em}``handling'': ``unproved'' / ``assumed'' / ``derived'' / ``absent'',\\
\hspace*{1em}``evidence\_in\_fl'': ``verbatim Lean code snippet supporting the judgment, or empty string'',\\
\hspace*{1em}``judgment'': ``faithful''/``reverted''/``other'',\\
\hspace*{1em}``reason'': ``one-sentence explanation''\}
\end{tcolorbox}

Two deterministic checks follow each judge call. First, the quoted evidence must occur in the FL output after comments are removed, ignoring whitespace; otherwise the output is labeled unclear. Second, an ``unproved'' call must quote an explicit placeholder or a \texttt{:= by} that no tactic follows; if the quoted code contains tactics, the claim is recorded as derived.

\section{Full Results}
\label{app:full-results}

\subsection{Detailed Metrics}
The main tables in \Cref{sec:exp-robustness,sec:exp-faithfulness} report only the main metric of each experiment, and we report the remaining metrics here. For global perturbations, \Cref{tab:exp1-full} reports TC, StmtSC, and ProofSC of every model under all five input styles on both datasets, in addition to FullyCorrect in \Cref{tab:robustness}. For local perturbations, \Cref{tab:edit-minif2f-full,tab:edit-math500-full} report TC, StmtSC, ProofSC, and FullyCorrect of every model under each perturbation type on miniF2F and MATH-500, respectively, in addition to FR and RR in \Cref{tab:proof-edit-results,tab:statement-edit-results}. For local perturbations, StmtSC and ProofSC are judged against the edited NL input, and TC is not a correctness target, since a faithful formalization of a counterfactual edit may not type-check. A faithful output can still pass TC in three situations. First, the edited input may remain valid: for example, a changed premise may still imply the conclusion. Second, the edit may make the premises contradictory, in which case any conclusion can be derived from them. Third, the output may leave the affected step as \texttt{sorry}, which TC accepts. Such placeholders rarely affect FullyCorrect: of the 1,902 FullyCorrect outputs of ProofBridge, ProofFlow, Kimina-Prover-RL-1.7B, and Kimina-Prover-Distill-8B under global perturbations (\Cref{tab:robustness}), 12 ($0.6\%$) contain \texttt{sorry} or \texttt{admit} outside comments, all of them from ProofFlow. The identifiable fraction (the complement of UR) is reported in \Cref{tab:identifiable-fraction}. Of the 413 step-deletion outputs labeled faithful, 58 have an empty proof body, 40 of them from Kimina-Prover-RL-1.7B and 17 from Kimina-Prover-Distill-8B. Without these outputs, the step-deletion FR on miniF2F and MATH-500 falls to $12.9\%$ and $39.6\%$ for Kimina-Prover-RL-1.7B and to $9.6\%$ and $32.4\%$ for Kimina-Prover-Distill-8B, changes by at most $0.7$ points for the other five models, and stays below $40\%$ for every model.

\label{app:nl-style-full-results}

\begin{table*}[t]
\centering

\setlength{\tabcolsep}{4pt}
\renewcommand{\arraystretch}{1.05}
\adjustbox{max width=.85\textwidth}{\begin{tabular}{lccccc|ccccc}
\toprule
\multirow{2}{*}{\textbf{Model}}
& \multicolumn{5}{c|}{\textbf{miniF2F}}
& \multicolumn{5}{c}{\textbf{MATH-500}} \\
\cmidrule(lr){2-6} \cmidrule(lr){7-11}
& Orig & G-FF & G-Step & Q-FF & Q-Step
& Orig & G-FF & G-Step & Q-FF & Q-Step \\
\midrule
Kimina-Prover-RL-1.7B
& 9.4 & 10.2 & 11.5 & 11.5 & 8.6
& 6.6 & 8.2 & 6.8 & 8.2 & 9.6 \\
Kimina-Prover-Distill-8B
& 19.3 & 16.0 & 16.4 & 19.7 & 16.0
& 19.8 & 16.4 & 16.8 & 16.8 & 19.0 \\
DeepSeek-Prover-V2-7B
& 1.2 & 8.6 & 7.4 & 12.7 & 8.6
& 8.2 & 7.4 & 5.8 & 6.4 & 5.0 \\
Goedel-Prover-V2-8B
& 1.2 & 2.5 & 4.1 & 3.7 & 4.1
& 4.4 & 2.0 & 2.0 & 3.8 & 3.0 \\
\midrule
Gemini-3.1-Pro
& 17.2 & {21.7} & {18.9} & {21.7} & {22.5}
& {26.4} & {23.6} & {24.6} & {23.8} & {23.8} \\
\midrule
ProofBridge
& {21.3} & 19.7 & 18.4 & 17.6 & 17.6
& 13.4 & 14.4 & 12.6 & 14.8 & 15.2 \\
ProofFlow
& 9.0 & 6.1 & 8.2 & 7.0 & 9.8
& 10.8 & 8.2 & 8.0 & 8.4 & 13.0 \\
\bottomrule
\end{tabular}}
\caption{Robustness of proof autoformalization models under global perturbations, measured by FullyCorrect (\%) on miniF2F and MATH-500 across five styles of natural-language theorem--proof inputs. An output is FullyCorrect only if it passes TC, StmtSC, and ProofSC. Models are grouped as in \Cref{tab:proof-edit-results}.}
\label{tab:robustness}
\end{table*}

\begin{table*}[!t]
\centering

\small
\setlength{\tabcolsep}{2.5pt}
\renewcommand{\arraystretch}{1.05}
\adjustbox{max width=\textwidth}{
\begin{tabular}{ll ccccc ccccc}
\toprule
\multirow{2}{*}{\textbf{Metric}} & \multirow{2}{*}{\textbf{Model}}
& \multicolumn{5}{c}{\textbf{miniF2F}} & \multicolumn{5}{c}{\textbf{MATH-500}} \\
\cmidrule(lr){3-7} \cmidrule(lr){8-12}
& & Orig & G-FF & G-Step & Q-FF & Q-Step & Orig & G-FF & G-Step & Q-FF & Q-Step \\
\midrule
\multirow{7}{*}{TC} & Kimina-Prover-RL-1.7B & 38.5 & 32.4 & 39.3 & 40.2 & 45.9 & 39.8 & 38.8 & 38.0 & 39.2 & 37.4 \\
& Kimina-Prover-Distill-8B & 44.7 & 39.3 & 40.6 & 43.4 & 46.7 & 49.2 & 48.2 & 44.6 & 46.8 & 50.0 \\
& DeepSeek-Prover-V2-7B & 5.7 & 17.2 & 19.3 & 28.7 & 27.5 & 25.8 & 25.4 & 22.0 & 26.0 & 20.8 \\
& Goedel-Prover-V2-8B & 9.8 & 11.9 & 12.3 & 14.8 & 15.2 & 16.8 & 16.4 & 14.6 & 17.8 & 18.4 \\
& Gemini-3.1-Pro & 46.3 & 50.4 & 49.2 & 47.1 & 53.7 & 58.0 & 56.8 & 57.8 & 55.6 & 54.4 \\
& ProofBridge & 37.3 & 36.5 & 37.3 & 39.3 & 38.9 & 33.2 & 35.0 & 35.2 & 31.6 & 33.6 \\
& ProofFlow & 26.6 & 19.7 & 26.6 & 25.4 & 26.2 & 27.4 & 21.0 & 23.2 & 22.6 & 32.4 \\
\cmidrule(lr){1-12}
\multirow{7}{*}{StmtSC} & Kimina-Prover-RL-1.7B & 38.1 & 39.3 & 42.6 & 43.9 & 41.4 & 29.2 & 32.4 & 28.8 & 31.6 & 28.0 \\
& Kimina-Prover-Distill-8B & 74.2 & 63.1 & 65.2 & 66.8 & 69.7 & 58.4 & 56.8 & 59.2 & 60.0 & 58.8 \\
& DeepSeek-Prover-V2-7B & 57.4 & 63.1 & 66.0 & 64.3 & 58.2 & 50.8 & 53.0 & 49.6 & 50.6 & 49.6 \\
& Goedel-Prover-V2-8B & 50.8 & 46.7 & 51.2 & 52.9 & 51.2 & 42.2 & 38.6 & 39.8 & 40.8 & 40.4 \\
& Gemini-3.1-Pro & 82.0 & 79.5 & 79.9 & 78.3 & 81.1 & 81.4 & 81.8 & 82.0 & 82.8 & 81.2 \\
& ProofBridge & 71.3 & 69.7 & 69.7 & 71.3 & 71.3 & 50.8 & 51.6 & 48.8 & 50.4 & 49.0 \\
& ProofFlow & 42.2 & 56.6 & 39.3 & 42.2 & 40.6 & 47.6 & 58.4 & 60.6 & 58.2 & 45.8 \\
\cmidrule(lr){1-12}
\multirow{7}{*}{ProofSC} & Kimina-Prover-RL-1.7B & 20.9 & 23.0 & 20.1 & 23.0 & 17.6 & 16.8 & 15.8 & 18.0 & 19.2 & 19.2 \\
& Kimina-Prover-Distill-8B & 41.4 & 36.9 & 38.5 & 38.5 & 40.2 & 43.6 & 39.4 & 38.2 & 40.2 & 39.0 \\
& DeepSeek-Prover-V2-7B & 40.2 & 41.4 & 38.5 & 40.2 & 43.0 & 43.2 & 41.4 & 43.6 & 43.6 & 43.4 \\
& Goedel-Prover-V2-8B & 27.0 & 27.9 & 34.0 & 34.0 & 33.6 & 29.4 & 27.4 & 29.0 & 28.4 & 25.8 \\
& Gemini-3.1-Pro & 36.5 & 38.1 & 37.7 & 39.8 & 36.1 & 41.6 & 40.4 & 41.8 & 41.2 & 39.8 \\
& ProofBridge & 38.1 & 43.9 & 41.0 & 37.3 & 43.0 & 37.0 & 41.8 & 43.2 & 39.6 & 45.2 \\
& ProofFlow & 31.1 & 36.5 & 33.6 & 30.7 & 36.1 & 34.2 & 34.8 & 38.0 & 39.8 & 33.8 \\
\bottomrule
\end{tabular}}
\caption{Global perturbation full results (\%).}
\label{tab:exp1-full}
\end{table*}

\label{app:edit-full-results}

\begin{table*}[!t]
\centering

\small
\setlength{\tabcolsep}{4pt}
\begin{tabular}{ll|cccc|cc}
\toprule
\textbf{Model} & \textbf{Perturbation} & \textbf{TC} & \textbf{StmtSC} & \textbf{ProofSC} & \textbf{FullyCorrect} & \textbf{FR} & \textbf{RR} \\
\midrule
\multirow{5}{*}{Kimina-Prover-RL-1.7B} & Number, statement & 23.8 & 14.1 & 7.5 & 2.2 & 82.9 & 17.1 \\
& Number, proof & 42.7 & 40.7 & 14.5 & 4.6 & 7.3 & 92.7 \\
& Symbol, statement & 16.2 & 23.1 & 4.6 & 0.8 & 90.4 & 9.6 \\
& Symbol, proof & 47.7 & 43.6 & 12.8 & 6.2 & 3.6 & 96.4 \\
& Step deletion & 31.8 & 34.9 & 14.7 & 4.7 & 42.6 & 57.4 \\
\cmidrule(lr){1-8}
\multirow{5}{*}{Kimina-Prover-Distill-8B} & Number, statement & 39.6 & 45.8 & 30.8 & 7.9 & 75.0 & 25.0 \\
& Number, proof & 41.9 & 65.1 & 29.0 & 12.0 & 5.3 & 94.7 \\
& Symbol, statement & 32.3 & 40.0 & 16.2 & 3.1 & 83.1 & 16.9 \\
& Symbol, proof & 46.7 & 69.7 & 29.7 & 12.8 & 8.3 & 91.7 \\
& Step deletion & 41.9 & 66.7 & 31.8 & 10.9 & 18.5 & 81.5 \\
\cmidrule(lr){1-8}
\multirow{5}{*}{DeepSeek-Prover-V2-7B} & Number, statement & 16.7 & 35.7 & 25.6 & 0.9 & 66.3 & 33.7 \\
& Number, proof & 18.3 & 63.5 & 32.0 & 5.0 & 8.3 & 91.7 \\
& Symbol, statement & 17.7 & 39.2 & 23.8 & 1.5 & 73.8 & 26.2 \\
& Symbol, proof & 21.0 & 67.2 & 26.2 & 4.6 & 7.9 & 92.1 \\
& Step deletion & 9.3 & 55.0 & 26.4 & 0.8 & 25.0 & 75.0 \\
\cmidrule(lr){1-8}
\multirow{5}{*}{Goedel-Prover-V2-8B} & Number, statement & 1.8 & 26.0 & 12.8 & 0.0 & 58.9 & 41.1 \\
& Number, proof & 13.3 & 49.0 & 21.6 & 1.2 & 10.1 & 89.9 \\
& Symbol, statement & 11.5 & 30.8 & 8.5 & 0.0 & 73.1 & 26.9 \\
& Symbol, proof & 4.6 & 49.7 & 23.1 & 0.5 & 2.6 & 97.4 \\
& Step deletion & 8.5 & 49.6 & 26.4 & 1.6 & 37.1 & 62.9 \\
\cmidrule(lr){1-8}
\multirow{5}{*}{Gemini-3.1-Pro} & Number, statement & 46.3 & 28.2 & 32.6 & 3.1 & 22.8 & 77.2 \\
& Number, proof & 48.5 & 81.3 & 29.5 & 17.0 & 4.1 & 95.9 \\
& Symbol, statement & 41.5 & 23.1 & 30.8 & 3.1 & 23.3 & 76.7 \\
& Symbol, proof & 45.6 & 76.4 & 26.2 & 15.4 & 8.3 & 91.7 \\
& Step deletion & 40.3 & 80.6 & 31.8 & 10.1 & 21.4 & 78.6 \\
\cmidrule(lr){1-8}
\multirow{5}{*}{ProofBridge} & Number, statement & 29.1 & 57.7 & 30.4 & 6.2 & 78.6 & 21.4 \\
& Number, proof & 36.1 & 71.8 & 29.5 & 12.9 & 39.4 & 60.6 \\
& Symbol, statement & 33.8 & 55.4 & 31.5 & 8.5 & 80.0 & 20.0 \\
& Symbol, proof & 37.9 & 65.6 & 25.6 & 10.8 & 44.9 & 55.1 \\
& Step deletion & 34.1 & 66.7 & 26.4 & 9.3 & 13.6 & 86.4 \\
\cmidrule(lr){1-8}
\multirow{5}{*}{ProofFlow} & Number, statement & 18.9 & 33.5 & 24.2 & 0.9 & 52.8 & 47.2 \\
& Number, proof & 17.4 & 54.8 & 27.4 & 5.4 & 32.6 & 67.4 \\
& Symbol, statement & 11.5 & 30.0 & 18.5 & 0.8 & 64.8 & 35.2 \\
& Symbol, proof & 19.5 & 60.0 & 26.2 & 7.2 & 27.0 & 73.0 \\
& Step deletion & 19.4 & 58.9 & 34.1 & 5.4 & 28.6 & 71.4 \\
\bottomrule
\end{tabular}
\caption{Local perturbation full results on miniF2F (\%).}
\label{tab:edit-minif2f-full}
\end{table*}

\begin{table*}[!t]
\centering

\small
\setlength{\tabcolsep}{4pt}
\begin{tabular}{ll|cccc|cc}
\toprule
\textbf{Model} & \textbf{Perturbation} & \textbf{TC} & \textbf{StmtSC} & \textbf{ProofSC} & \textbf{FullyCorrect} & \textbf{FR} & \textbf{RR} \\
\midrule
\multirow{5}{*}{Kimina-Prover-RL-1.7B} & Number, statement & 24.9 & 8.8 & 7.3 & 0.6 & 62.4 & 37.6 \\
& Number, proof & 35.6 & 25.2 & 12.3 & 3.9 & 5.2 & 94.8 \\
& Symbol, statement & 27.2 & 9.7 & 4.6 & 0.5 & 82.8 & 17.2 \\
& Symbol, proof & 39.3 & 23.5 & 11.2 & 4.6 & 8.1 & 91.9 \\
& Step deletion & 34.9 & 28.6 & 19.3 & 8.3 & 59.7 & 40.3 \\
\cmidrule(lr){1-8}
\multirow{5}{*}{Kimina-Prover-Distill-8B} & Number, statement & 34.5 & 23.3 & 29.0 & 1.8 & 49.1 & 50.9 \\
& Number, proof & 45.0 & 54.4 & 33.5 & 11.0 & 2.9 & 97.1 \\
& Symbol, statement & 30.3 & 28.2 & 21.0 & 3.1 & 67.2 & 32.8 \\
& Symbol, proof & 45.0 & 52.4 & 34.1 & 13.5 & 3.1 & 96.9 \\
& Step deletion & 40.6 & 53.6 & 37.5 & 14.6 & 37.8 & 62.2 \\
\cmidrule(lr){1-8}
\multirow{5}{*}{DeepSeek-Prover-V2-7B} & Number, statement & 22.2 & 21.0 & 32.7 & 1.8 & 48.1 & 51.9 \\
& Number, proof & 25.8 & 49.7 & 35.0 & 4.3 & 6.4 & 93.6 \\
& Symbol, statement & 14.9 & 30.3 & 24.6 & 1.5 & 69.4 & 30.6 \\
& Symbol, proof & 35.8 & 49.6 & 35.2 & 7.4 & 8.5 & 91.5 \\
& Step deletion & 25.0 & 43.2 & 39.1 & 5.7 & 35.2 & 64.8 \\
\cmidrule(lr){1-8}
\multirow{5}{*}{Goedel-Prover-V2-8B} & Number, statement & 18.8 & 13.9 & 17.6 & 0.2 & 44.6 & 55.4 \\
& Number, proof & 3.9 & 38.4 & 20.4 & 0.2 & 6.6 & 93.4 \\
& Symbol, statement & 9.7 & 22.1 & 12.8 & 0.5 & 59.3 & 40.7 \\
& Symbol, proof & 16.9 & 33.0 & 18.6 & 2.0 & 9.2 & 90.8 \\
& Step deletion & 14.1 & 33.3 & 19.3 & 3.1 & 39.5 & 60.5 \\
\cmidrule(lr){1-8}
\multirow{5}{*}{Gemini-3.1-Pro} & Number, statement & 63.7 & 19.8 & 39.2 & 3.5 & 14.2 & 85.8 \\
& Number, proof & 57.9 & 81.6 & 32.9 & 19.0 & 6.3 & 93.7 \\
& Symbol, statement & 47.2 & 22.1 & 31.3 & 2.6 & 26.9 & 73.1 \\
& Symbol, proof & 49.3 & 79.9 & 33.8 & 18.9 & 4.1 & 95.9 \\
& Step deletion & 56.2 & 81.2 & 40.1 & 22.4 & 36.5 & 63.5 \\
\cmidrule(lr){1-8}
\multirow{5}{*}{ProofBridge} & Number, statement & 33.3 & 29.6 & 39.6 & 4.1 & 57.4 & 42.6 \\
& Number, proof & 33.9 & 44.4 & 38.4 & 12.1 & 41.4 & 58.6 \\
& Symbol, statement & 26.7 & 32.3 & 32.3 & 4.1 & 70.8 & 29.2 \\
& Symbol, proof & 27.5 & 44.4 & 35.0 & 9.7 & 41.4 & 58.6 \\
& Step deletion & 37.5 & 40.6 & 37.0 & 9.9 & 27.7 & 72.3 \\
\cmidrule(lr){1-8}
\multirow{5}{*}{ProofFlow} & Number, statement & 16.9 & 17.3 & 28.4 & 0.6 & 49.3 & 50.7 \\
& Number, proof & 19.6 & 53.4 & 29.4 & 5.7 & 24.4 & 75.6 \\
& Symbol, statement & 11.3 & 28.7 & 21.5 & 0.5 & 57.7 & 42.3 \\
& Symbol, proof & 19.5 & 56.4 & 25.5 & 6.3 & 20.2 & 79.8 \\
& Step deletion & 22.4 & 58.9 & 37.0 & 6.8 & 32.9 & 67.1 \\
\bottomrule
\end{tabular}
\caption{Local perturbation full results on MATH-500 (\%).}
\label{tab:edit-math500-full}
\end{table*}

\begin{table*}[!t]
\centering

\setlength{\tabcolsep}{5pt}
\renewcommand{\arraystretch}{1.05}
\adjustbox{max width=.7\textwidth}{\begin{tabular}{l|cc|cc|cc|cc}
\toprule
\multirow{3}{*}{\textbf{Model}}
& \multicolumn{4}{c|}{\textbf{Number edit}}
& \multicolumn{4}{c}{\textbf{Symbol edit}} \\
\cmidrule(lr){2-5} \cmidrule(lr){6-9}
& \multicolumn{2}{c|}{miniF2F} & \multicolumn{2}{c|}{MATH-500}
& \multicolumn{2}{c|}{miniF2F} & \multicolumn{2}{c}{MATH-500} \\
\cmidrule(lr){2-3} \cmidrule(lr){4-5}
\cmidrule(lr){6-7} \cmidrule(lr){8-9}
& FR & RR
& FR & RR
& FR & RR
& FR & RR \\
\midrule
Kimina-Prover-RL-1.7B
& {82.9} & {17.1}
& {62.4} & {37.6}
& {90.4} & {9.6}
& {82.8} & {17.2} \\
Kimina-Prover-Distill-8B
& 75.0 & 25.0
& 49.1 & 50.9
& 83.1 & 16.9
& 67.2 & 32.8 \\
DeepSeek-Prover-V2-7B
& 66.3 & 33.7
& 48.1 & 51.9
& 73.8 & 26.2
& 69.4 & 30.6 \\
Goedel-Prover-V2-8B
& 58.9 & 41.1
& 44.6 & 55.4
& 73.1 & 26.9
& 59.3 & 40.7 \\
\midrule
Gemini-3.1-Pro
& 22.8 & 77.2
& 14.2 & 85.8
& 23.3 & 76.7
& 26.9 & 73.1 \\
\midrule
ProofBridge
& 78.6 & 21.4
& 57.4 & 42.6
& 80.0 & 20.0
& 70.8 & 29.2 \\
ProofFlow
& 52.8 & 47.2
& 49.3 & 50.7
& 64.8 & 35.2
& 57.7 & 42.3 \\
\bottomrule
\end{tabular}}
\caption{Faithful Rate (FR) and Reverted Rate (RR) (\%) for statement-level local perturbations, computed over identifiable outputs (faithful or reverted). Unclear outputs are excluded, and the identifiable fraction is reported in \Cref{tab:identifiable-fraction}. Models are grouped as in \Cref{tab:proof-edit-results}.}
\label{tab:statement-edit-results}
\end{table*}

\subsection{Composition of the Unclear Outputs}
\label{app:other-breakdown}

This section analyzes the unclear outputs (\Cref{sec:faithfulness-metrics}), i.e., the outputs we cannot label faithful or reverted. Inspecting these outputs, we find several recurring patterns. A small portion are degenerate, where the targeted region contains no theorem at all, or the output is empty or an error. The rest are genuine formalizations in which the edited value never appears at the target position, neither the new value nor the original one.
This is most common in proof edits, where the proof is closed by an automation tactic such as \texttt{norm\_num}, \texttt{ring}, or \texttt{linarith} that never writes the value out, or reaches the goal by a different route. For instance, in the proof edit that changes $49$ to $48$ in \texttt{amc12a\_2020\_p7}, one model's entire proof is \texttt{:= by norm\_num}, so neither value appears and we label the output unclear. Step deletion behaves the same way: the deleted step's result usually appears nowhere in the FL proof. In short, an unclear label means that the edit cannot be located in the output, not that the choice between faithful and reverted was difficult.

\begin{figure*}[t]
    \centering
    \includegraphics[width=\textwidth]{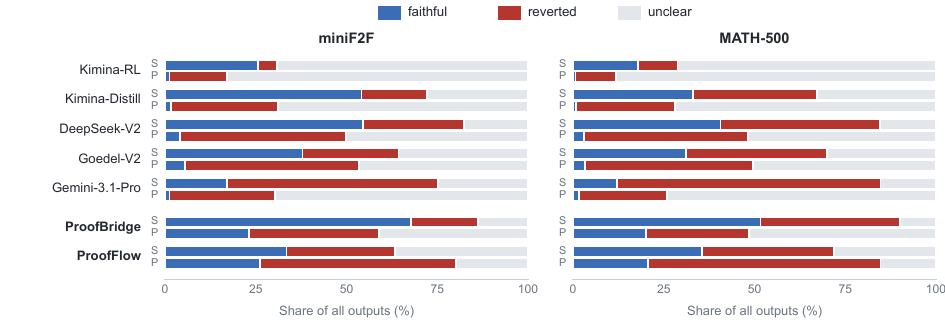}
    \caption{Outcome of number edits as a share of all outputs, obtained by combining FR and RR (\Cref{tab:proof-edit-results,tab:statement-edit-results}) with the identifiable fraction (\Cref{tab:identifiable-fraction}). S: the edit is in the statement; P: the edit is in the proof.}
    \label{fig:edit-outcomes}
\end{figure*}

\Cref{tab:identifiable-fraction} reports the identifiable fraction ($1$ minus UR) for each model and perturbation type. It is typically highest for statement edits and lowest for proof-level number and symbol edits, where an automation tactic usually closes the proof without writing the edited value; step deletion mostly lies in between. \Cref{fig:edit-outcomes} shows the same outcome for number edits as shares of all outputs: counted this way, the five non-specialized models follow a proof-level number edit in fewer than $6\%$ of all outputs, so the conclusion that proof-level edits are rarely followed holds whether or not unclear outputs are excluded. \Cref{tab:fr-intervals} further gives the counts behind the proof-level FR with 95\% Wilson intervals~\citep{wilson1927probable}; for the five non-specialized models, the upper end of every interval on number and symbol edits is below $22\%$.

\begin{table*}[!t]
\centering

\small
\setlength{\tabcolsep}{2.5pt}
\resizebox{\textwidth}{!}{
\begin{tabular}{l|cccc|cccc|cc}
\toprule
\multirow{2}{*}{\textbf{Model}}
& \multicolumn{4}{c|}{\textbf{Statement edits}}
& \multicolumn{4}{c|}{\textbf{Proof edits}}
& \multicolumn{2}{c}{\textbf{Step deletion}} \\
& Num$_{\text{mF2F}}$ & Num$_{\text{M500}}$ & Sym$_{\text{mF2F}}$ & Sym$_{\text{M500}}$
& Num$_{\text{mF2F}}$ & Num$_{\text{M500}}$ & Sym$_{\text{mF2F}}$ & Sym$_{\text{M500}}$
& mF2F & M500 \\
\midrule
Kimina-Prover-RL-1.7B & 30.8 & 28.8 & 40.0 & 29.7 & 17.0 & 11.9 & 14.4 & 10.6 & 36.4 & 37.5 \\
Kimina-Prover-Distill-8B & 72.2 & 67.3 & 63.8 & 59.5 & 31.1 & 28.2 & 24.6 & 27.5 & 62.8 & 57.8 \\
DeepSeek-Prover-V2-7B & 82.4 & 84.5 & 82.3 & 82.1 & 49.8 & 48.3 & 39.0 & 40.4 & 68.2 & 56.3 \\
Goedel-Prover-V2-8B & 64.3 & 70.0 & 71.5 & 63.1 & 53.5 & 49.5 & 39.5 & 37.2 & 48.1 & 44.8 \\
Gemini-3.1-Pro & 75.3 & 84.7 & 69.2 & 74.4 & 30.3 & 26.0 & 18.5 & 20.9 & 54.3 & 59.9 \\
ProofBridge & 86.3 & 90.0 & 88.5 & 86.2 & 58.9 & 48.5 & 45.6 & 41.5 & 62.8 & 58.3 \\
ProofFlow & 63.4 & 72.0 & 67.7 & 66.7 & 80.1 & 84.7 & 75.9 & 70.8 & 76.0 & 77.6 \\
\bottomrule
\end{tabular}}
\caption{Identifiable fraction (\%): the share of outputs classified as faithful or reverted, over which FR/RR are computed (the complement of the Unclear Rate, UR).}
\label{tab:identifiable-fraction}
\end{table*}

\begin{table*}[t]
\centering
\footnotesize
\adjustbox{max width=\textwidth}{\begin{tabular}{lcccccc}
\toprule
\multirow{2}{*}{\textbf{Model}} & \multicolumn{2}{c}{\textbf{Number edit}} & \multicolumn{2}{c}{\textbf{Symbol edit}} & \multicolumn{2}{c}{\textbf{Step deletion}} \\
\cmidrule(lr){2-3} \cmidrule(lr){4-5} \cmidrule(lr){6-7}
& miniF2F & MATH-500 & miniF2F & MATH-500 & miniF2F & MATH-500 \\
\midrule
Kimina-Prover-RL-1.7B & 3/41 [2.5, 19.4] & 3/58 [1.8, 14.1] & 1/28 [0.6, 17.7] & 3/37 [2.8, 21.3] & 20/47 [29.5, 56.7] & 43/72 [48.2, 70.3] \\
Kimina-Prover-Distill-8B & 4/75 [2.1, 12.9] & 4/138 [1.1, 7.2] & 4/48 [3.3, 19.6] & 3/96 [1.1, 8.8] & 15/81 [11.6, 28.3] & 42/111 [29.4, 47.1] \\
DeepSeek-Prover-V2-7B & 10/120 [4.6, 14.7] & 15/236 [3.9, 10.2] & 6/76 [3.7, 16.2] & 12/141 [4.9, 14.3] & 22/88 [17.1, 35.0] & 38/108 [26.8, 44.6] \\
Goedel-Prover-V2-8B & 13/129 [6.0, 16.5] & 16/242 [4.1, 10.5] & 2/77 [0.7, 9.0] & 12/130 [5.4, 15.4] & 23/62 [26.2, 49.5] & 34/86 [29.9, 50.1] \\
\midrule
Gemini-3.1-Pro & 3/73 [1.4, 11.4] & 8/127 [3.2, 11.9] & 3/36 [2.9, 21.8] & 3/73 [1.4, 11.4] & 15/70 [13.4, 32.4] & 42/115 [28.3, 45.6] \\
\midrule
ProofBridge & 56/142 [31.8, 47.7] & 98/237 [35.3, 47.7] & 40/89 [35.0, 55.3] & 60/145 [33.7, 49.5] & 11/81 [7.8, 22.7] & 31/112 [20.2, 36.6] \\
ProofFlow & 63/193 [26.4, 39.5] & 101/414 [20.5, 28.8] & 40/148 [20.5, 34.7] & 50/247 [15.7, 25.7] & 28/98 [20.6, 38.2] & 49/149 [25.9, 40.8] \\
\bottomrule
\end{tabular}}
\caption{Faithful outputs over identifiable outputs for proof-level local perturbations, with the 95\% Wilson interval of the Faithful Rate (FR, \%) in brackets. The counts are recovered from the rates in \Cref{tab:proof-edit-results,tab:identifiable-fraction} and the numbers of edits in \Cref{tab:edit-counts}. Models are grouped as in \Cref{tab:proof-edit-results}.}
\label{tab:fr-intervals}
\end{table*}

\section{Additional Results}
\label{app:additional-results}

This section tests the sensitivity of our results to several experimental choices, i.e., whether our conclusions hold across proof lengths and when multiple candidates are sampled and selected with the Lean type checker. It also measures how much of the variation across styles comes from sampling alone.

\label{app:additional-results}

\paragraph{Proof length.}
To check whether the sensitivity to input styles is limited to proofs of specific length ranges, we compare problems with shorter and longer NL proofs.
We count the whitespace-delimited words in the original NL proof of each problem,
and place the problem in the longer group if its proof has at least 100 words and in the
shorter group otherwise. Each problem stays in its group across all five styles.
For each group and each of the two specialized proof autoformalization models (ProofBridge and ProofFlow), we compute FullyCorrect and ProofSC for each style
and report the range (maximum minus minimum, in pp) across styles
(\Cref{tab:additional-proof-length}). The ranges are substantial in both groups, and neither group
is consistently more sensitive than the other. Thus, the sensitivity to input styles exists
for proofs of different lengths.

\begin{table*}[!t]
\centering

\small
\begin{tabular}{llrrrrrr}
\toprule
& & \multicolumn{2}{c}{Problems} & \multicolumn{2}{c}{FullyCorrect range} & \multicolumn{2}{c}{ProofSC range} \\
Dataset & Model & Shorter & Longer & Shorter & Longer & Shorter & Longer \\
\midrule
miniF2F & ProofBridge & 181 & 63 & 6.1 & 6.3 & 8.3 & 17.5 \\
miniF2F & ProofFlow & 181 & 63 & 3.3 & 6.3 & 6.6 & 6.3 \\
MATH-500 & ProofBridge & 350 & 150 & 4.3 & 4.7 & 9.4 & 5.3 \\
MATH-500 & ProofFlow & 350 & 150 & 6.3 & 5.3 & 5.7 & 10.0 \\
\bottomrule
\end{tabular}
\caption{Proof-length analysis for ProofBridge and ProofFlow.
Shorter/longer: fewer than 100 / at least 100 words in the original NL proof.
Ranges (pp) are taken across the five input styles.}
\label{tab:additional-proof-length}
\end{table*}

\paragraph{Verifier-guided selection.}
Our main experiments use a single output per input. Here we test whether sampling multiple candidates and selecting one with the Lean type checker reduces the sensitivity to input styles.
We evaluate ProofBridge, the strongest specialized proof autoformalization model in our evaluation (\Cref{tab:robustness}),
with $k$ candidates sampled for each input and report $k=1$ and $k=8$.
For each problem and style, we select the first of the $k$ candidates, in sampling
order, that passes TC. If no candidate passes TC, the input is counted as unsuccessful.
Otherwise, StmtSC and ProofSC are judged on the selected candidate, and FullyCorrect
requires TC, StmtSC, and ProofSC on this same candidate. Rates are computed over all
inputs and averaged over the five input styles.
We measure the sensitivity to input styles by the FullyCorrect range across styles and by the flip rate:
for each problem, we compare all ten pairs of the five styles, count a pair as flipped if one
selected output is FullyCorrect and the other is not, and average the fraction of flipped pairs over problems.
As shown in \Cref{tab:additional-verifier-selection}, from $k=1$ to $8$, TC and FullyCorrect increase substantially on both datasets, while the flip rate also increases on both datasets, and the FullyCorrect range increases on miniF2F and remains similar on MATH-500.
Thus, verifier-guided selection improves the absolute success rate but does not
eliminate the inconsistency across styles.

\begin{table*}[!t]
\centering

\small
\setlength{\tabcolsep}{4pt}
\adjustbox{max width=\textwidth}{
\begin{tabular}{lrrrr}
\toprule
Dataset & TC & FullyCorrect & FullyCorrect range & Flip rate \\
\midrule
miniF2F & $37.9\to87.3$ & $18.9\to33.9$ & $3.7\to8.6$ & $20.2\to24.7$ \\
MATH-500 & $33.7\to80.8$ & $14.1\to27.1$ & $2.6\to2.2$ & $14.9\to21.0$ \\
\bottomrule
\end{tabular}}
\caption{Verifier-guided selection for ProofBridge with $k=1\to8$ candidates. TC and FullyCorrect (\%) are averaged over the five input styles; the FullyCorrect range across styles is in pp, and the flip rate is in \%.}
\label{tab:additional-verifier-selection}
\end{table*}

\paragraph{Sampling variation.}
Our main experiments use one sampled output per input, so part of the difference between two styles comes from sampling alone. To measure this part, we sample eight outputs per input for three models (ProofBridge, Kimina-Prover-RL-1.7B, and Kimina-Prover-Distill-8B) and compare how often two outputs disagree on FullyCorrect when they come from the same input and when they come from two styles of the same problem. For ProofBridge on miniF2F, for example, the two rates are $18.4\%$ and $20.1\%$. Disagreement is higher across styles than within a style for all three models on both datasets, by $0.2$--$1.7$ pp, and the 95\% bootstrap interval of this difference (resampling problems) excludes zero in five of the six cases (all except Kimina-Prover-RL-1.7B on miniF2F); when only StmtSC and ProofSC are considered, it excludes zero in all six. The input style thus has a consistent effect on top of sampling, which accounts for most of the disagreement between two single outputs.

\section{Reliability Validation}
\label{app:reliability-detail}

This section details the reliability checks summarized in \Cref{sec:reliability-data}. We check three things: that the perturbations are constructed correctly, that the semantic-correctness judge agrees with human judgment, and that the faithfulness judge reports trustworthy verdicts.

\paragraph{Perturbation reliability.}
We verify that every edit is exactly the intended one. For number and symbol edits we compare the original and edited NL inputs character by character: the recorded span must hold the old value in the original and the new value in the edited string, and the text on both sides of the span must match exactly, so the change is localized with no other edits. For example, in \texttt{aime\_1983\_p1} the span covers the token \texttt{40} in the fragment $\log_y w = 40$; we confirm the original holds \texttt{40} at that span and the edited string holds \texttt{43}, while the $112$ characters before the span and every character after it match exactly, so \texttt{40}$\,\to\,$\texttt{43} is the only difference. Symbol edits are checked the same way, for instance a single $+ \to -$ that turns $x^2 + 18x + 30$ into $x^2 - 18x + 30$. For step deletions, we locate the changed region directly as the gap between the longest common prefix and the longest common suffix of the two proofs, and require the edited proof to be strictly shorter and to differ in a single contiguous span; in \texttt{aime\_1983\_p3}, this isolates one removed sentence of $151$ characters lying between a $300$-character shared prefix and a $359$-character shared suffix. All 2,637 local edits pass these checks. For global perturbations, an automated check finds that $96.4\%$ of the 2,976 perturbed inputs preserve the exact set of numeric and symbolic tokens of the original; the rest are formatting differences such as digit grouping or numbers spelled out, not content changes, so this figure is a lower bound.

\paragraph{Semantic-correctness judge.}
We check the StmtSC and ProofSC judge against independent human annotation. For each dimension, an author labels $100$ judge verdicts as correct or not, reading each against the NL theorem or proof. We deliberately balance each sample, $50$ verdicts the judge accepted and $50$ it rejected, so that it is tested on both the cases it accepts and the cases it rejects. The judge matches the human label on $94$ of the $100$ StmtSC cases and $90$ of the $100$ ProofSC cases, with Cohen's $\kappa$~\citep{cohen1960coefficient} of $0.88$ and $0.80$, confirming that its verdicts are reliable on this balanced sample. We further compare the judge with a second judge in \Cref{sec:reliability-data}.

\paragraph{Faithfulness judge.}
We check the FR/RR/UR judge in three ways. First, the judge reports the value it reads from the FL output, which lets us test each verdict for internal consistency without a second model: a \texttt{faithful} verdict should carry the edited value, a \texttt{reverted} verdict the original value, and an \texttt{other} verdict neither, with equivalent forms such as $6.5$ and $13/2$ treated as equal. The verdict and the reported value agree in $96.7\%$ of the 16,212 number and symbol cases from all seven models.
Second, we require the judge to copy the verbatim snippet of the FL output supporting its verdict and confirm that this snippet occurs in the scored fragment; this check passes on $99.8\%$ of cases. Checking that the evidence stated by a model is grounded in its input has also been used to support model answers with verified quotes~\citep{menick2022teaching} and to audit language model classifiers~\citep{liu2026explanations}.
Third, an author manually checks $100$ number and symbol verdicts and $50$ step-deletion verdicts against the model output and the judge's evidence field, flagging only two of the $150$ as potentially incorrect. For step deletion, the two deterministic checks in \Cref{app:edit-judge-prompts} also apply to every label: of the 1,344 labels that quote evidence, the quoted code occurs in the comment-free output for 1,280 ($95.2\%$), and the remaining 64 are labeled unclear.

\begin{table}[t]
\centering
\small
\setlength{\tabcolsep}{4pt}
\begin{tabular}{lrrrr}
\toprule
Metric & Gemini & GPT & Agree. & Cohen's $\kappa$ \\
\midrule
StmtSC & 67.5 & 69.0 & 88.5 & 0.735 \\
ProofSC & 48.5 & 51.0 & 84.5 & 0.690 \\
\bottomrule
\end{tabular}
\caption{Positive rates (\%) of StmtSC and ProofSC judged by Gemini-2.5-Flash and GPT-5.4-mini on the same 200 outputs, together with the label agreement (\%) and Cohen's $\kappa$ between the two judges.}
\label{tab:additional-cross-judge}
\end{table}

\section{Potential Risks and Use of AI Assistants}
\label{app:risk-ai}

\paragraph{Potential risks.} Our work focuses on mathematical problems and poses no foreseeable risk.

\paragraph{Use of AI assistants.} AI assistants were used to assist with coding, to polish the writing, and to help organize the paper, with some code logic drawing on their suggestions.

\end{document}